\documentclass[twoside]{article}

\usepackage[accepted]{aistats2022}
\usepackage[round]{natbib}

\usepackage{multirow}
\usepackage[utf8]{inputenc}
\usepackage{amsmath,amssymb,array}
\usepackage{bbm}
\usepackage{xcolor}
\usepackage{graphicx}
\usepackage{xargs}
\usepackage{algorithmic}
\usepackage[font=small,skip=0pt]{caption}
\usepackage{wrapfig}
\usepackage{caption}

\usepackage{multicol}
\usepackage{booktabs}
\usepackage{todonotes}
\usepackage[american]{babel}
\usepackage{natbib}

\renewcommand\citet{\cite}
\usepackage[hyphens]{url}
\usepackage{hyperref}
\usepackage{colortbl}
\usepackage{csquotes} 
\usepackage{mathtools}
\usepackage{float} 
\usepackage{adjustbox}
\usepackage{dsfont}
\usepackage[ruled, vlined, linesnumbered]{algorithm2e}
\let\oldnl\nl
\newcommand{\nonl}{\renewcommand{\nl}{\let\nl\oldnl}}
\usepackage{cleveref}
\usepackage{autonum}
\newcommand{\Xspace}{\mathcal{X}}                                           
\newcommand{\Yspace}{\mathcal{Y}}                                           
\newcommand{\D}{\mathcal{D}}                                                
\newcommand{\Dset}{\left( \left(\mathbf{x}^{(1)}, y^{(1)}\right), \ldots, \left(\mathbf{x}^{(n)},  y^{(n)}\right)\right)}    
\renewcommand{\xi}[1][i]{\mathbf{x}^{(#1)}}                                          
\newcommand{\yi}[1][i]{y^{(#1)}}                                            
\newcommand{\xivec}{\left(x^{(i)}_1, \ldots, x^{(i)}_p\right)^T}            
\newcommand{\xj}{\xv_j}                                                       
\newcommand{\xjvec}{\left(x^{(1)}_j, \ldots, x^{(n)}_j\right)^T}            

\newcommand{\fh}{\hat{f}}                                                   
\ifdefined\N                                                                
\renewcommand{\N}{\mathds{N}}                                                
\else
  \newcommand{\N}{\mathds{N}}
\fi
\ifdefined\C
  \renewcommand{\C}{\mathds{C}}                                             
\else
  \newcommand{\C}{\mathds{C}}
\fi

\def\argmin{\mathop{\sf arg\,min}}                                          

\newcommand{\xv}{\mathbf{x}}													

\renewcommand{\P}{\mathds{P}}                                               

\newcommand{\fS}{f_S}                                           
\newcommand{\fhS}{\fh_S}                                        

\renewcommand{\Dset}{\{ (\xi, \yi) \}_{i=1}^n}
\newtheorem{theorem}{Theorem}
\newtheorem{corollary}{Corollary}[theorem]

\begin{document}

\twocolumn[

\aistatstitle{REPID: Regional Effect Plots with implicit Interaction Detection}

\aistatsauthor{ Julia Herbinger \And Bernd Bischl \And  Giuseppe Casalicchio }

\aistatsaddress{LMU Munich \And LMU Munich \And LMU Munich } ]

\begin{abstract}
Machine learning models can automatically learn complex relationships, such as non-linear and interaction effects.
Interpretable machine learning methods such as partial dependence plots visualize marginal feature effects but may lead to misleading interpretations when feature interactions are present. 
Hence, employing additional methods that can detect and measure the strength of interactions is paramount to better understand the inner workings of machine learning models. 
We demonstrate several drawbacks of existing global interaction detection approaches, characterize them theoretically, and evaluate them empirically.
Furthermore, we introduce \textit{regional effect plots with implicit interaction detection}, a novel framework to detect interactions between a feature of interest and other features.  
The framework also quantifies the strength of interactions and provides interpretable and distinct regions in which feature effects can be interpreted more reliably, as they are less confounded by interactions.
We prove the theoretical eligibility of our method and show its applicability on various simulation and real-world examples.  
\end{abstract}

\section{INTRODUCTION}

Many machine learning (ML) models are considered black-boxes, as they do not provide insights into how the model's prediction function is composed and which features or interactions\footnote{Interactions describe to what extent a feature's effect on the model prediction is influenced by other features.} are actually used by the model.
This lack of transparency has been partially addressed by recent developments in the field of interpretable ML.
In general, the literature distinguishes between local and global interpretation methods \citep{molnar2020interpretable}.
Global interpretation methods aim at explaining the overall behavior of an ML model. 
Examples include the partial dependence (PD) plot \citep{friedman2001greedy}, which visualizes the effect of a feature on the model's prediction, and the permutation feature importance, which quantifies the relevance of features
\citep{fisher2019all}. 
However, many of these global interpretation methods are confounded by feature interactions, meaning that they can produce misleading explanations when feature interactions are present
because they often aggregate over individual effects of local interpretation methods and thereby obfuscate heterogeneous effects induced by feature interactions \citep{molnar2021general}.
This so-called \textit{aggregation bias} \citep{mehrabi2021survey} is responsible for producing global explanations that are usually not representative or not valid for many individuals.
Instead of explaining the ML model on a global level, local interpretation methods -- such as individual conditional expectation (ICE) curves \citep{goldstein2015peeking}, LIME \citep{ribeiro2016should}, or Shapley values \citep{strumbelj2014} -- can be used to understand how a feature influences an individual prediction. 
However, many local interpretation methods do not provide a global understanding of the ML model due to their local view (i.e., their explanations only refer to individual observations). 
Thus, it is often recommended to consider both local and global interpretation methods.
For example, in the case of PD plots, looking additionally at ICE curves \citep{goldstein2015peeking} can help to reveal interactions when the ICE curves are heterogeneous (see Figure \ref{fig:sim1_allcurves}).
Yet, ICE curves are not able to quantify the strength of the underlying feature interactions, nor can they tell exactly which features interact with each other. 
On the other hand, other methods that quantify the interaction strength between features are available. However, they do not provide any visual component of how these interactions influence the effect of a feature of interest \citep{friedman2008predictive, greenwell2018simple}.
The work in this paper is motivated by subgroup analysis \citep{su2009subgroup} as a trade-off between local and global explanations.
We aim to uncover a possible \textit{aggregation bias} in the PD plot by finding interpretable subgroups in the data with differing influences of a feature on the predictions. Hence, for well-performing ML models, this might also reveal a possible bias in the data (e.g., when the influence of a feature on the prediction strongly differs for certain subgroups, although it should not) and thus might be helpful to uncover possible negative societal impacts.

\textbf{Contributions:}
We introduce \textit{regional effect plots with implicit interaction detection} (REPID), a model-agnostic interpretation method that produces regional effect plots (REPs) in which feature effects are less confounded by interactions. 
Regions are obtained by a decision tree and thus represent interpretable and distinct subgroups in the feature space.
We also propose a new measure to detect and quantify interactions with a feature of interest, which can be used to rank interactions according to their strength. 
To receive a broader and more competitive comparability, we derive another global interaction index based on SHAP interaction values \citep{lundberg2018consistent}.
We mathematically prove the theoretical meaningfulness of our method and demonstrate its advantages compared to not only the well-known H-statistic \citep{friedman2008predictive}, but also to Greenwell's interaction index \citep{greenwell2018simple} and our derived global SHAP interaction index.
Finally, we demonstrate the usefulness of our method on real-world data.

\textbf{Open Science:} 
The implementation of the proposed method and the fully reproducible code for all experiments are provided in a public repository\footnote{\url{https://github.com/JuliaHerbinger/repid}}. 






\section{BACKGROUND AND RELATED WORK}
\label{sec:rel_work}

\emph{Notation:} Consider a $p-$dimensional feature space $\Xspace \in \mathbb{R}^p$ and a target space $\Yspace$ (e.g., $\Yspace = \mathbb{R}$ for regression).
The corresponding random variables are $X = (X_1, \hdots, X_p)$ for the features and $Y$ for the target.
ML algorithms learn a prediction model $\fh$ 
using training data $\D = \Dset$ sampled i.i.d. from the unknown joint distribution $\P_{X,Y}$.
In our notation, the $i$-th observation is denoted by $\xi = \xivec$, and $\xj = \xjvec$ denotes the realizations of the $j$-th feature $X_j$.

\emph{PD Plot \citep{friedman2001greedy}:} The marginal relationship of features on model predictions can be visualized by PD plots.
Consider a set of feature indices $S \subseteq \{1, \hdots, p\}$ and its complement $C = S^\complement$.
Each observation $\xi$ can be partitioned into $\xi_S$ and $\xi_C$ containing only features indexed by $S$ and $C$, respectively.
$X_S$ and $X_C$ refer to the corresponding random variables. 
The PD function of features indexed by $S$ marginalizes over features in $C$ and is defined as 
$\fS ^{PD}(\xv_S) = E_{X_C} [\fh(\xv_S, X_C) ]$. 
The PD function is estimated by Monte-Carlo integration:
\begin{equation}
\label{eq:pdp}
\textstyle \fhS^{PD}(\xv_S) = \frac{1}{n} \sum_{i = 1}^n \fh(\xv_S, \xi_C).
\end{equation}
Here, $\fh(\xv_S, \xi_C)$ can be read as the prediction of the $i$-th observation where features in $S$ were replaced by $\xv_S$. 
Plotting the pairs $\{ (\xv_S^{(k)}, \fhS(\xv_S^{(k)}))\}_{k=1}^m$ using grid points\footnote{Common choices are randomly selected feature values, quantiles, or equidistant values \citep{molnar2021general}.} denoted by $\xv_S^{(1)}, \hdots, \xv_S^{(m)}$ yields a PD curve.
The mean-centered PD function can be estimated by
\begin{equation}
\label{eq:cpdp}
\textstyle \fhS^{PD,c}(\xv_S) = \fhS^{PD}(\xv_S) - \frac{1}{m} \sum_{k = 1}^m \fh_S^{PD}(\xv_S^{(k)}).
\end{equation}
If $|S| = 2$, we get a 2-dimensional PD plot showing the joint marginal effect of the 2 features included in $S$.

\emph{ICE Plot \citep{goldstein2015peeking}:} The averaging in Eq.~\eqref{eq:pdp} can obfuscate complex relationships resulting from feature interactions. 
ICE plots address this problem by directly visualizing individual curves for each observation, i.e., $\{ (\xv_S^{(k)}, \fh(\xv_S^{(k)}, \xi_C) ) \}_{k=1}^m$ for all $i \in \{1, \hdots, n\}$.
ICE curves will usually have different shapes if interactions with other features in $C$ are present.
To facilitate the visual identification of heterogeneous ICE curves and, consequently, the presence of interactions, the authors propose the derivative-ICE (d-ICE) plot.
Assuming that there are no interactions between features $\xv_S$ and $\xv_C$, the prediction function can be written as $\hat{f}(\mathbf{x}) = \hat{f}(\xv_S, \xv_C) = g(\xv_S) + h(\xv_C)$.
Hence, the partial derivatives of all ICE curves $\frac{\delta \fh(\xv_S, \xi_C)}{\delta \xv_S} = g'(\xv_S)$ do not depend on $\xi_C$, which means that d-ICE curves have the same shape if there are no interactions.
The d-ICE plot visualizes the partial derivatives of ICE curves
along with their standard deviation to highlight regions in $\xv_S$ where the d-ICE curves are heterogeneous (see Figure \ref{fig:sim1_allcurves}).

\emph{Visual INteraction Effects (VINE) \citep{britton2019vine}:}
The principle of VINE is to cluster similar slopes of ICE curves to obtain clusters where the curves are less affected by interactions based on a three-step approach: 
(1) for a feature of interest, find clusters where the ICE curves of that feature have similar slopes using, e.g., agglomerative clustering, 
(2) for each found cluster, create a binary label containing the information of whether an observation belongs to the considered cluster or any other cluster and apply a tree stump, 
(3) identify the split feature and its split point and merge clusters that use the same feature and a similar split point.
Although VINE is based on a similar strategy as our approach, its three-step approach has several disadvantages (see Section \ref{sec:rep_motiv}). Approaches to group ICE curves to reduce feature dependencies instead of feature interactions is introduced in \cite{molnar2021modelagnostic} and \cite{groemping:2020}.

\emph{H-Statistic \citep{friedman2008predictive}:}
The H-Statistic is based on the assumption that if two features do not interact, the 2-dimensional mean-centered PD function of two features $\xv_j$ and $\xv_l$ is additively separable and can be decomposed into the sum of their mean-centered 1-dimensional PDs, i.e.,
\begin{equation}
\label{eq:interaction}
f_S^{PD,c}(\xv_S) = f^{PD,c}_{j}(\xv_j) + f^{PD,c}_{l}(\xv_l), \text{ with } S = \{j,l\}. 
\end{equation}
The stronger an interaction effect, the more the sum of $f^{PD,c}_{j}(\xv_j)$ and $f^{PD,c}_{l}(\xv_l)$ will deviate from $f_S^{PD,c}(\xv_S)$.
Hence, the H-statistic computes the interaction strength between two features $\xv_j$ and $\xv_l$ by quantifying the degree of this deviation using
\begin{equation} \textstyle
    \mathcal{\hat H}_{S}^2 = \frac{\sum\nolimits_{i=1}^n \left( 
    \hat f_{S}^{PD,c}(\xi_S) -  \sum\nolimits_{k \in S} \hat f_{k}^{PD,c}(\xi_k) 
    \right)^2}{\sum_{i=1}^n \left( \hat f_{S}^{PD,c}(\xi_S) \right)^2 }.
\label{eq:hstatistic}
\end{equation}

\emph{Greenwell's interaction index \citep{greenwell2018simple}:} The interaction strength between two features $\xv_j$ and $\xv_l$ is quantified based on the variability of the PD function of $\xv_j$ conditioned on a fixed value of $\xv_l$ (see Appendix \ref{app:proofs_main}). 

However, the H-Statistic and the Greenwell's interaction index only quantify interaction effects and do not visualize how interactions influence the marginal effect of a feature. Moreover, both methods are sensitive to varying main effects (see Section \ref{sec:quant_motiv} and \ref{sec:sim_weak}). 

\emph{Functional ANOVA (fANOVA) \citep{hooker2004discovering}:}
The fANOVA decomposes the prediction function as follows:
\begin{equation}
\hat{f}(\xv) = 
\textstyle g_0 + \sum_{k = 1}^p \; \sum_{W \subseteq \{1,\ldots,p\}, |W| = k} g_W(\xv_W)
\label{eq:fANOVA}
\end{equation}
where $E_X[g_W(\xv_W)] = 0$ for all feature index sets $W$ (zero-means property).
While $g_W(X_W)$ with $|W| = 1$ refers to main (or \textit{first-order}) effects, $g_W(X_W)$ with $|W| > 1$ refers to interactions (or \textit{higher-order}) effects.  
Based on the decomposition in Eq.~\eqref{eq:fANOVA}, the authors detect interactions of any order by applying an efficient search algorithm and visualize them in an interaction network graph. 
However, the network only shows the presence of feature interactions and does not quantify the interaction strength or illustrate how they influence the prediction.
A discussion on the assumptions and application of the fANOVA decomposition in the context of this paper is provided in Appendix \ref{app:proofs_general}.


\emph{SHAP interaction values \citep{lundberg2018consistent}:}
The method is based on Shapley values \citep{shapley1953value} and Shapley interaction indices \citep{fujimoto2006axiomatic} from game theory.
In the ML context, SHAP interaction values of two features quantify the pure interaction effect after accounting for the individual feature effects.
The SHAP interaction values separate the interaction effect from the main effects of two features indexed by $j$ and $l$ (for $j \neq l$) for an observation $\xv$: 
\begin{equation}
    \textstyle \Phi_{j,l}(\xv) = \sum_{ S \subseteq \{1, \dots p \} \setminus \{j,l\} } \frac{|S|! (p-|S|-2)!}{2(p-1)!} \nabla_{j,l}(\xv_S), 
\end{equation}
where $\nabla_{j,l}(\xv_S) = f_{S \cup \{j,l\}}^{PD}(\xv_{S \cup \{j,l\}}) - f_{S \cup \{j\}}^{PD}(\xv_{S \cup \{j\}}) - f_{S \cup \{l\}}^{PD}(\xv_{S \cup \{l\}}) + \fS^{PD}(\xv_{S})$.
The SHAP interaction values have only been introduced on an observational level, where the final plot over all observations shows the influence of the interaction effect on the prediction. 

\section{THE REPID METHOD}
\label{sec:pd_and_ice}
REPID visualizes regional marginal effects of a certain feature of interest $\xv_S$ with $|S| = 1$ depending on its interactions with other features and quantifies the underlying interaction strength. 
The following simulation example demonstrates the benefits of our method compared to existing ones. We draw $n = 500$ samples for 6 independent random variables, which are distributed as follows:
$X_1, X_2 \sim \mathcal{U}(-1,1)$, $X_3, X_5 \sim \mathcal{B}(n, 0.5)$, $X_4 \sim \mathcal{B}(n, 0.7)$ and $X_6 \sim \mathcal{N}(1,5)$. The true relationship is described by
\begin{equation}
f(\xv) = 0.2 \xv_1 - 8 \xv_2 + 8 \xv_2  \mathds{1}_{(\xv_1 > 0)} + 16 \xv_2  \mathds{1}_{(\xv_3 = 0)} + \epsilon
\label{eq:simexample}
\end{equation}
with $\epsilon \sim \mathcal{N}(0,1)$. 
We fit a random forest (RF) with 500 trees on the data. Due to the linear relationship, we can assume that the interaction strength between $\xv_2$ and $\xv_3$ is higher than the one between $\xv_2$ and $\xv_1$. 

\subsection{Regional Effect Plots}
\label{sec:rep}
\subsubsection{Motivation}
\label{sec:rep_motiv}
\begin{figure}[tbh]
    \centering
    \includegraphics[width=0.49\linewidth]{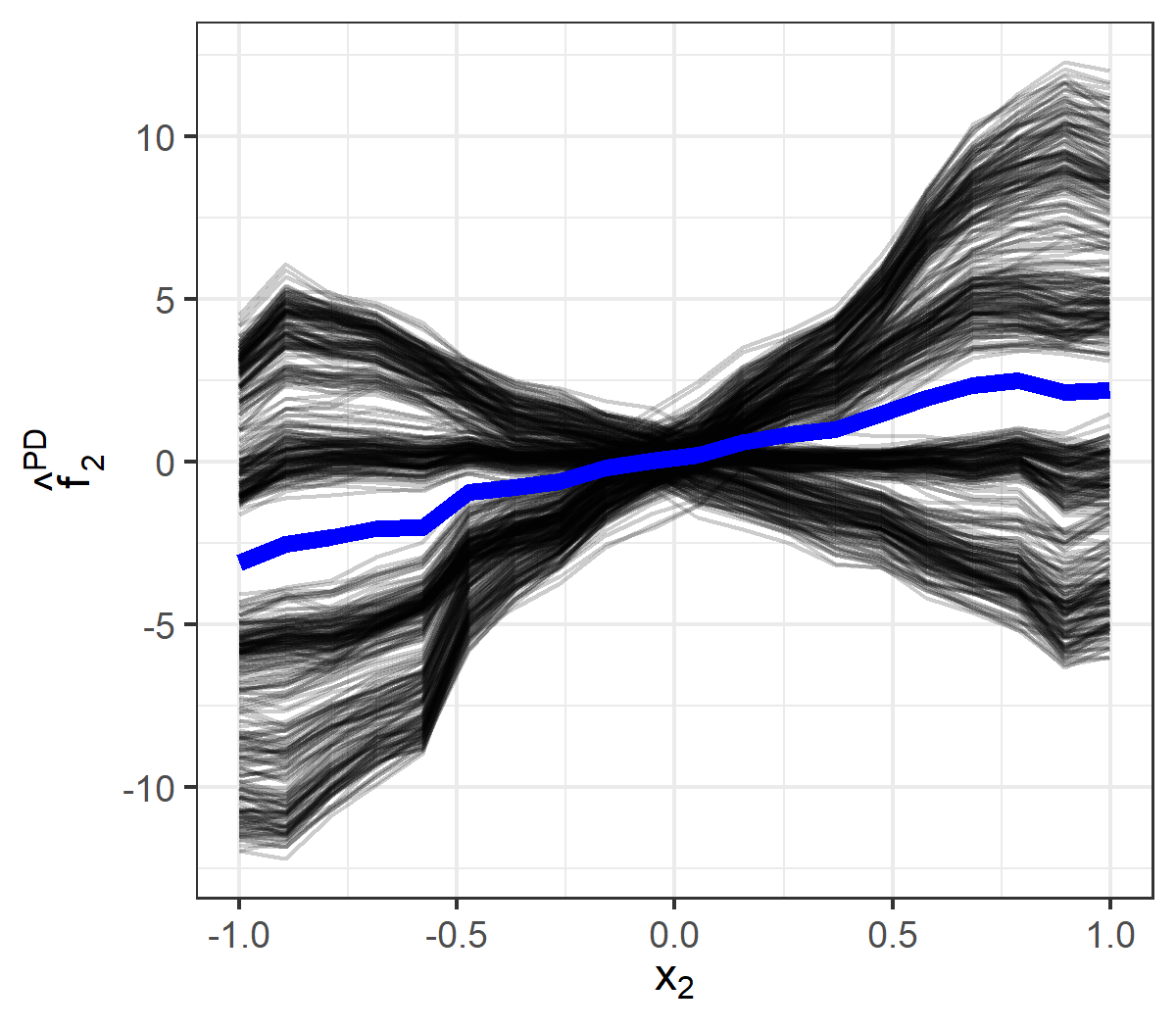}
    \includegraphics[width=0.49\linewidth]{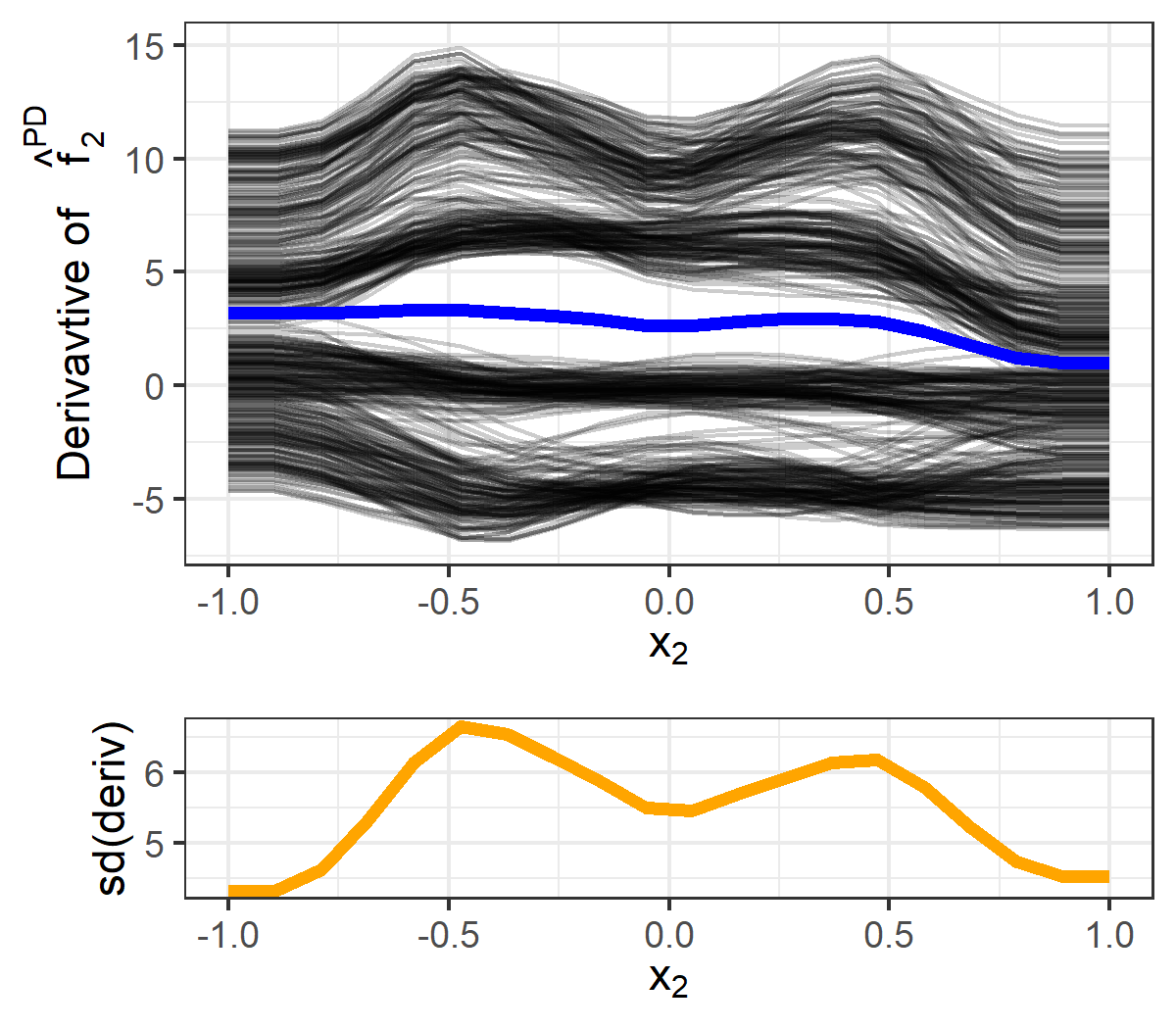}
    \vspace{.2in}
    \caption{Left: ICE curves (black) and PD plot (blue) for $\xv_2$. Right: Smoothed d-ICE curves (upper plot) and standard deviation of d-ICE curves (lower plot).}
    \label{fig:sim1_allcurves}
\end{figure}

PD plots are often shown together with their underlying ICE curves (see Figure \ref{fig:sim1_allcurves}). 
The heterogeneous shapes of ICE curves imply the presence of feature interactions.
Although ICE or d-ICE plots indicate interactions, they do not provide any information on which other features are responsible for these interactions and how the underlying interaction influences the marginal effect of $\xv_S$ (see Figure \ref{fig:sim1_allcurves}). 
Grouping homogeneous ICE curves will reduce the presence of individual interaction effects within a group. 
This leads to regional PD plots that actually reflect the pure marginal effect of $\xv_S$ within this group. 
VINE \citep{britton2019vine} implements this idea by clustering ICE curves with similar slopes (see Section \ref{sec:rel_work}). However, VINE is only a visual tool and does not quantify or rank feature interactions.
Furthermore, VINE is an unsupervised approach, and its solution depends on the number of clusters $k$ that must be chosen (which is not trivial). 
Another drawback is that VINE ``finds'' feature interactions in an inconvenient second step by fitting a separate tree stump for each cluster (see Section \ref{sec:rel_work}). 
Due to the different tree stumps used in VINE, the derived decision rules are often not distinct and therefore difficult to interpret.
In a third step, VINE introduces a post-hoc merging of clusters based on similar decision rules. 
In Figure 2, we show that this three-step approach does not always lead to meaningful groupings.
While in the left plot, the ICE curves are divided meaningfully into 2 clusters based on the most interacting feature $\xv_3$ (according to Eq.~\eqref{eq:simexample}), the clusters in the right plot do not divide the ICE curves into visually meaningful groups with homogeneous ICE curves. 
\begin{figure}[tbh]
    \centering
    \includegraphics[width=0.49\linewidth]{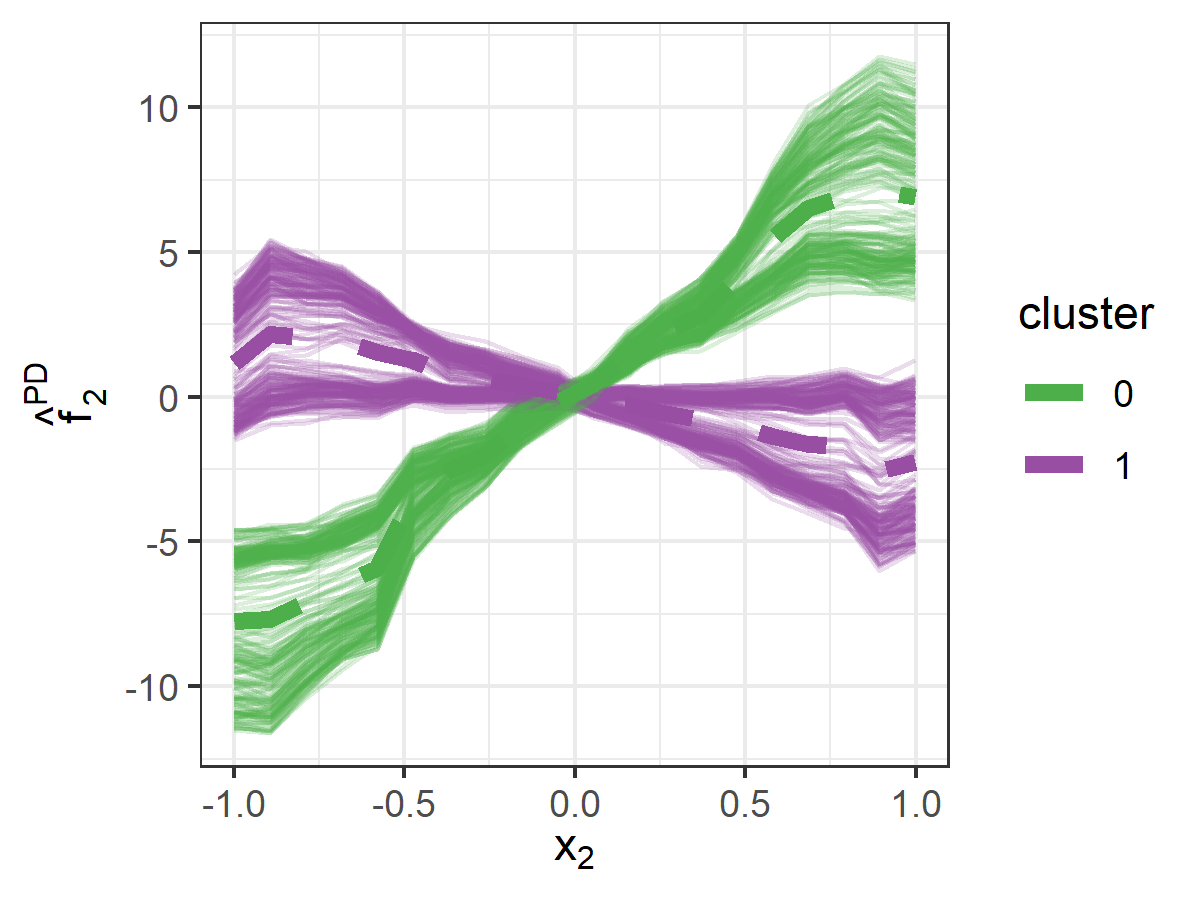}
    \includegraphics[width=0.49\linewidth]{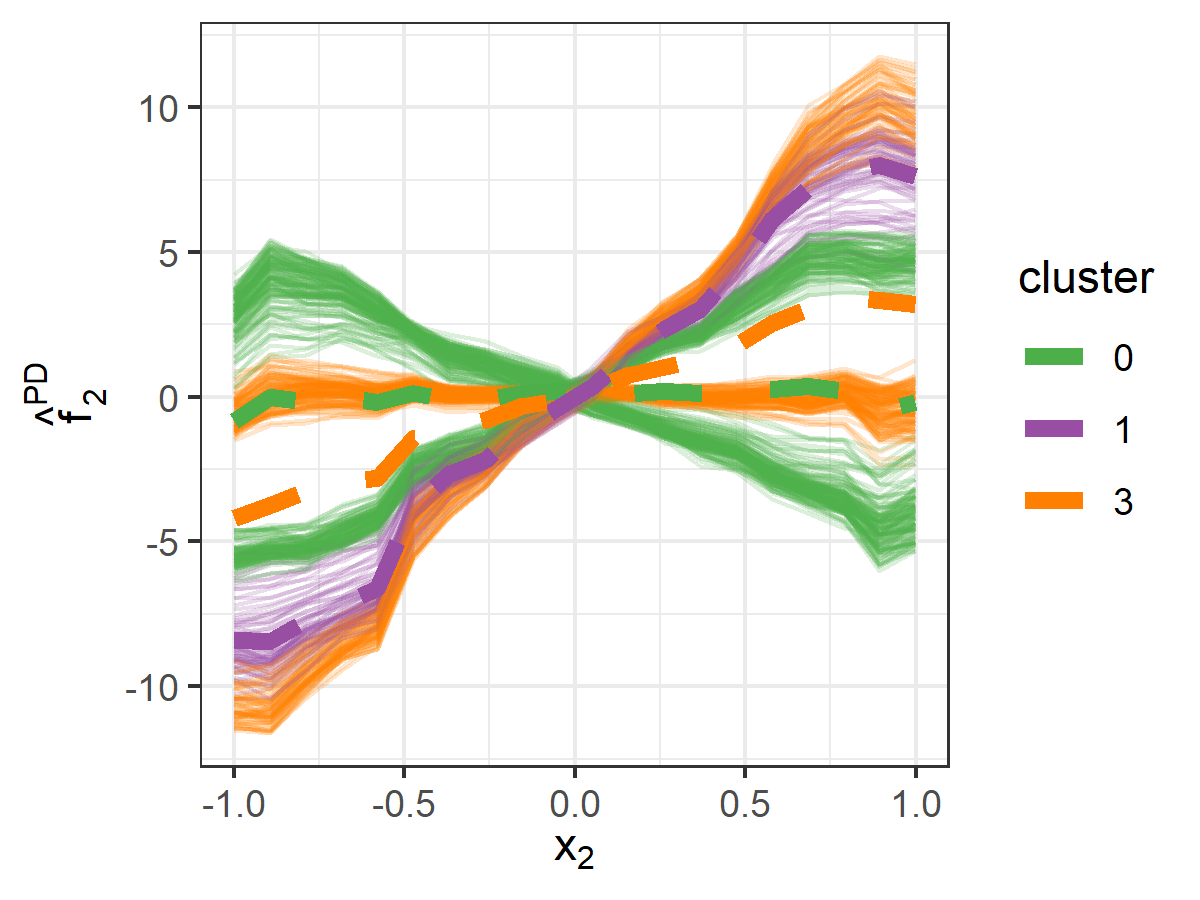}
    \vspace{.2in}
    \caption{ICE and regional PD (dashed) plot of $\xv_2$ clustered by VINE for $k = 2$ (left) and $k=5$ (right). The 5 clusters are reduced to 3 by post-hoc merging.
    Cluster numbers 0 and 3 still contain differing individual interaction effects, which are averaged and hence not represented well by the regional PD plot.
    }
    \label{fig:sim1_vine}
\end{figure}

\subsubsection{Methodology}
\label{sec:rep_method}
Here, we derive a new tree-based approach to determine optimal REPs for any feature of interest $\xv_S$.
REPs are regional PD plots that aggregate ICE curves within automatically identified regions where feature effects are less confounded by interactions. 
Our aim is to recursively split the entire data referred to by index set $\mathcal{N} = \{1, \dots, n\}$ into interpretable regions to obtain more homogeneous ICE curves for $\xv_S$ within the split regions denoted by $\mathcal{N}_g$ (where $g \in \{1, \ldots G\}$ indexes a certain node of the tree and $G$ is the number of all tree nodes). Hence, we want to split $\mathcal{N}$ in such a way that ICE curves within the obtained regions have a similar shape, meaning that the distance of these ICE curves to the REP estimate (i.e., $\hat{f}^{PD}_{S|\mathcal{N}_g} (\xv_S):= \frac{1}{|\mathcal{N}_g|}\sum_{i \in \mathcal{N}_g} \hat{f}\left(\xv_S, \xv_C^{(i)}\right)$) is small.
To that end, we propose a tree-based partitioning in Algorithm \ref{alg:tree}, which refers only to a single binary split and is inspired by the CART algorithm \citep{breiman:1984}\footnote{Algorithm \ref{alg:tree} is defined for numerical features. For categorical features, we use an exhaustive search as seen in CART. The computational feasibility of this procedure depends on the number of categories.}. 
The splitting is recursively repeated until the split criterion (denoted by $\mathcal{I}(\hat{t},\hat{j})$ in Algorithm \ref{alg:tree}) does not improve anymore compared to the previous split or until a pre-specified stop criterion is met. 
The split criterion is based on a suitable risk function $\mathcal{R}$ that operates on ICE curves (see also Eq.~\eqref{eq:risk_L2}).

  \begin{algorithm}[htb]
    \caption{Tree-based Partitioning}
    \label{alg:tree}
    \begin{algorithmic}
      \STATE \textbf{input: } index set $\mathcal{N}$, risk $\mathcal{R}_{L2}$ (see, e.g., Eq.~\eqref{eq:risk_L2})
      \STATE \textbf{output: } child nodes $\mathcal{N}_l^{\hat{t},\hat{j}}$ and $\mathcal{N}_r^{\hat{t}, \hat{j}}$
      \FOR{each feature indexed by $j \in C$}{
        \FOR{every split $t$ on feature $\xv_j$ }{
            \STATE $\mathcal{N}_l^{t,j} = \{ i \in \mathcal{N}\}_{\xv_j^{(i)} \leq t}$ ;
            $\mathcal{N}_r^{t,j} = \{i \in \mathcal{N}\}_{\xv_j^{(i)} > t}$ 
            \STATE $\mathcal{I}(t,j) = \mathcal{R}_{L2}(\mathcal{N}_l^{t,j}) + \mathcal{R}_{L2}(\mathcal{N}_r^{t,j})$
        }\ENDFOR
      } \ENDFOR   
      \STATE Choose $\hat{t}, \hat{j} \in \argmin\nolimits_{t,j} \mathcal{I}(t,j)$
    \end{algorithmic}
  \end{algorithm}

We first estimate the mean-centered ICE curves by $\hat{f}^c(\xv_S, \xv_C^{(i)}) = \hat{f}(\xv_S, \xv_C^{(i)}) - \frac{1}{m} \sum\nolimits_{k=1}^m \hat{f}(\xv_S^{(k)}, \xv_C^{(i)})$. 
Since we want to minimize the shape differences between ICE curves in the regions, we then define the risk function $\mathcal{R}_{L2}$ in Eq.~\eqref{eq:rl2}\footnote{Multiplying with $\frac{1}{m}$ to obtain the average loss can be neglected for optimization.} such that the variance (L2 loss) of the mean-centered ICE curves is minimized. 
This can be estimated by calculating the L2 loss of the mean-centered ICE curves at each grid point (see Eq.~\eqref{eq:splitting_crit_L2}) and aggregating it over all grid points:
\begin{eqnarray} \textstyle
    &\mathcal{L}\left(\mathcal{N}_g, x_S\right) =
    \sum\limits_{i \in \mathcal{N}_g}
     \left(\hat f^{c}(x_S, \xv_C^{(i)}) - \hat f_{S|\mathcal{N}_g}^{PD,c} (x_S)\right)^2
    \label{eq:splitting_crit_L2} \\
\label{eq:rl2}
    &\mathcal{R}_{L2}\left(\mathcal{N}_g\right) =
    \sum\limits_{k = 1}^m \mathcal{L}\left(\mathcal{N}_g, x_S^{(k)}\right) 
    \label{eq:risk_L2}
\end{eqnarray}



\begin{theorem}
If Eq.~\eqref{eq:fANOVA} holds, then 
$\hat f^c(\xv_S, \xi_C)$ with $|S| = 1$ can be decomposed into the
mean-centered\footnote{$g_W^{c_S}(X_W) = g_W(X_W) - E_{X_S}[g_W(X_W)]$ is the \textit{mean-centered} counterpart of $g_W(X_W)$ of Eq.~\eqref{eq:fANOVA} regarding $X_S$. 
} main effect of $\xv_S$ (i.e. $g^{c_S}_S(\xv_S)$) and the mean-centered interaction effect of $\xv_S$ with $\xv_{C}$ for the i-th observation (i.e., $g^{c_S}_{{C_k} \cup \{S\}}(\xv_S, \xi_{C_k})$): 
\begin{equation}
\textstyle \hat f^c(\xv_S, \xi_C) =  g^{c_S}_S(\xv_S) +  \sum\limits_{k = 1}^{p-1}  \sum\limits_{\substack{C_k\subseteq C, \\ |C_k| = k}} g^{c_S}_{{C_k} \cup \{S\}}(\xv_S, \xi_{C_k}).
\end{equation}

\label{theorem1}
\end{theorem}
%
\begin{corollary}
If Eq.~\eqref{eq:fANOVA} holds, then $f^{PD, c}_S(\xv_S) = E_{X_C}[\hat f^c(\xv_S, X_C) ]$ with $|S| = 1$ can be decomposed into 
\begin{equation}
\textstyle g^{c_S}_S(\xv_S) + \sum\limits_{k = 1}^{p-1}  \sum\limits_{\substack{C_k\subseteq C, \\ |C_k| = k}} E_{X_C}\left[g^{c_S}_{{C_k} \cup \{S\}}(\xv_S, X_{C_k})\right].
\end{equation}

The proof can be found in Appendix \ref{app:proof_t1}.
\label{corollary1}
\end{corollary}


Based on Theorem \ref{theorem1} and Corollary \ref{corollary1} -- where we show that the mean-centered ICE curves and PD function can be decomposed in first-order and higher-order terms which depend on $\xv_S$ -- we can prove in Theorem \ref{theorem2}, that our risk function of Eq.~\eqref{eq:risk_L2} only depends on the interaction effects between $\xv_S$ and features in $\xv_C$. Hence, by minimizing this risk function, we minimize the individual interaction effects between the feature of interest and all other features. Thus, we minimize the shape differences between ICE curves in each region.  
Theorem \ref{theorem3} states that the theoretical minimum of our split criterion leads to the optimal solution we aim to achieve, meaning that for each final region, all ICE curves are best represented by the REP.

\begin{theorem}
The distance minimized by the risk function $\mathcal{R}_{L2}$ of Eq.~\eqref{eq:risk_L2}
only depends on the mean-centered interaction effects between $\xv_S$ with $|S| = 1$ and all features interacting with $\xv_S$, i.e., for the i-th observation, the distance results in 
 \begin{equation} \textstyle
\sum\limits_{k = 1}^{p-1} \; \sum\limits_{\substack{C_k \subseteq C,\\ |C_k| = k}} g^{c_S}_{{C_k} \cup \{S\}}(\xv_S, \xv_{C_k}^{(i)}) - E_{X_{C}} [ g^{c_S}_{{C_k} \cup \{S\}}(\xv_S, X_{C_k})].
\end{equation}
The proof can be found in Appendix \ref{app:proof_t2}.
\label{theorem2}
\end{theorem}



\begin{figure}[tbh]
    \centering
      \includegraphics[width=0.49\linewidth]{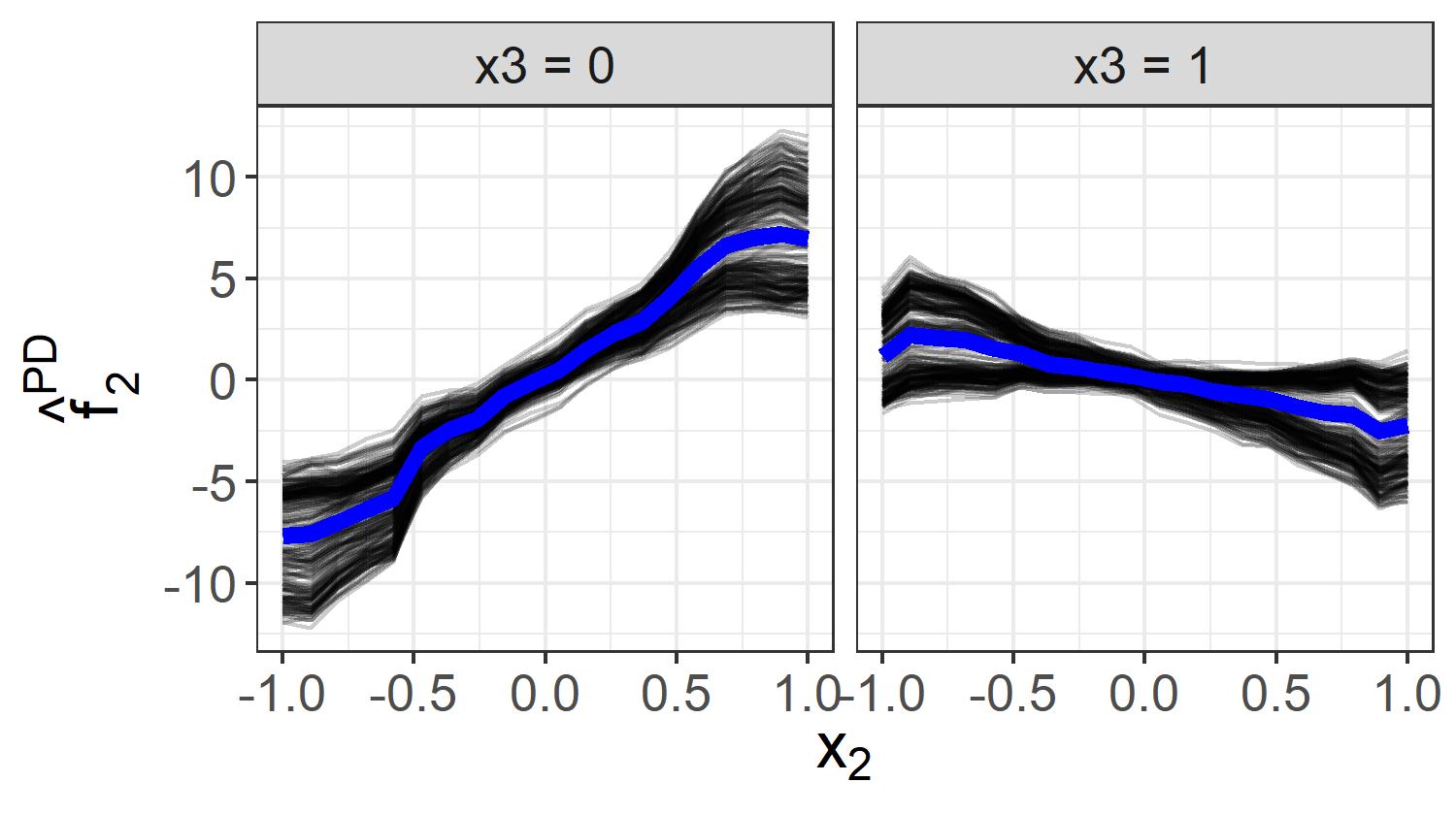}
      \scalebox{1}{
      \hspace{15pt} 
      \begin{tikzpicture}
      \usetikzlibrary{arrows}
        \usetikzlibrary{shapes}
         \tikzset{treenode/.style={draw}}
         \tikzset{line/.style={draw, thick}}
        \node [treenode](a0) {} ; [below=1pt,at=(4,0)]  {};
         \node [treenode, below=0.3cm, at=(a0.south), xshift=-2.0cm]  (a1) {};
         \node [treenode, below=0.3cm, at=(a0.south), xshift=-0.4cm]  (a2) {};
         \path [line] (a0.south) -- + (0,-0.2cm) -| (a1.north) node [midway, above] {};
         \path [line] (a0.south) -- +(0,-0.2cm) -|  (a2.north) node [midway, above] {};
      \end{tikzpicture}
      \hspace{35pt}
      \begin{tikzpicture}
      \usetikzlibrary{arrows}
        \usetikzlibrary{shapes}
         \tikzset{treenode/.style={draw}}
         \tikzset{line/.style={draw, thick}}
        \node [treenode] (a01) {};[below=5pt,at=(node1.south) , xshift=3.5cm]
         \node [treenode, below=0.3cm, at=(a01.south), xshift=0.4cm]  (a1) {};
         \node [treenode, below=0.3cm, at=(a01.south), xshift=2.0cm]  (a2) {};
         \path [line] (a01.south) -- + (0,-0.2cm) -| (a1.north) node [midway, above] {};
         \path [line] (a01.south) -- +(0,-0.2cm) -|  (a2.north) node [midway, above] {};
      \end{tikzpicture}
      }
    \includegraphics[width=0.49\linewidth]{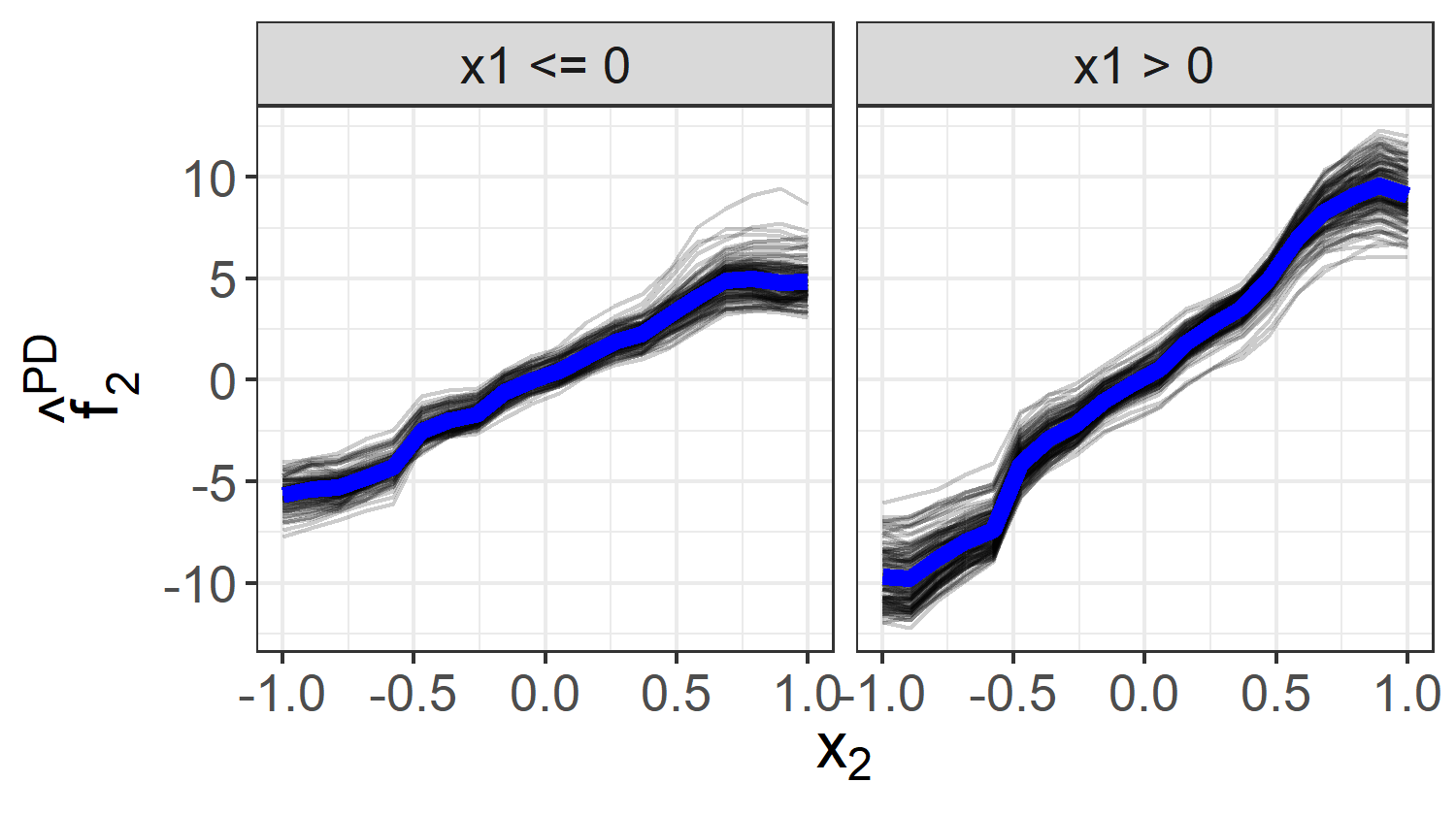}
    \includegraphics[width=0.49\linewidth]{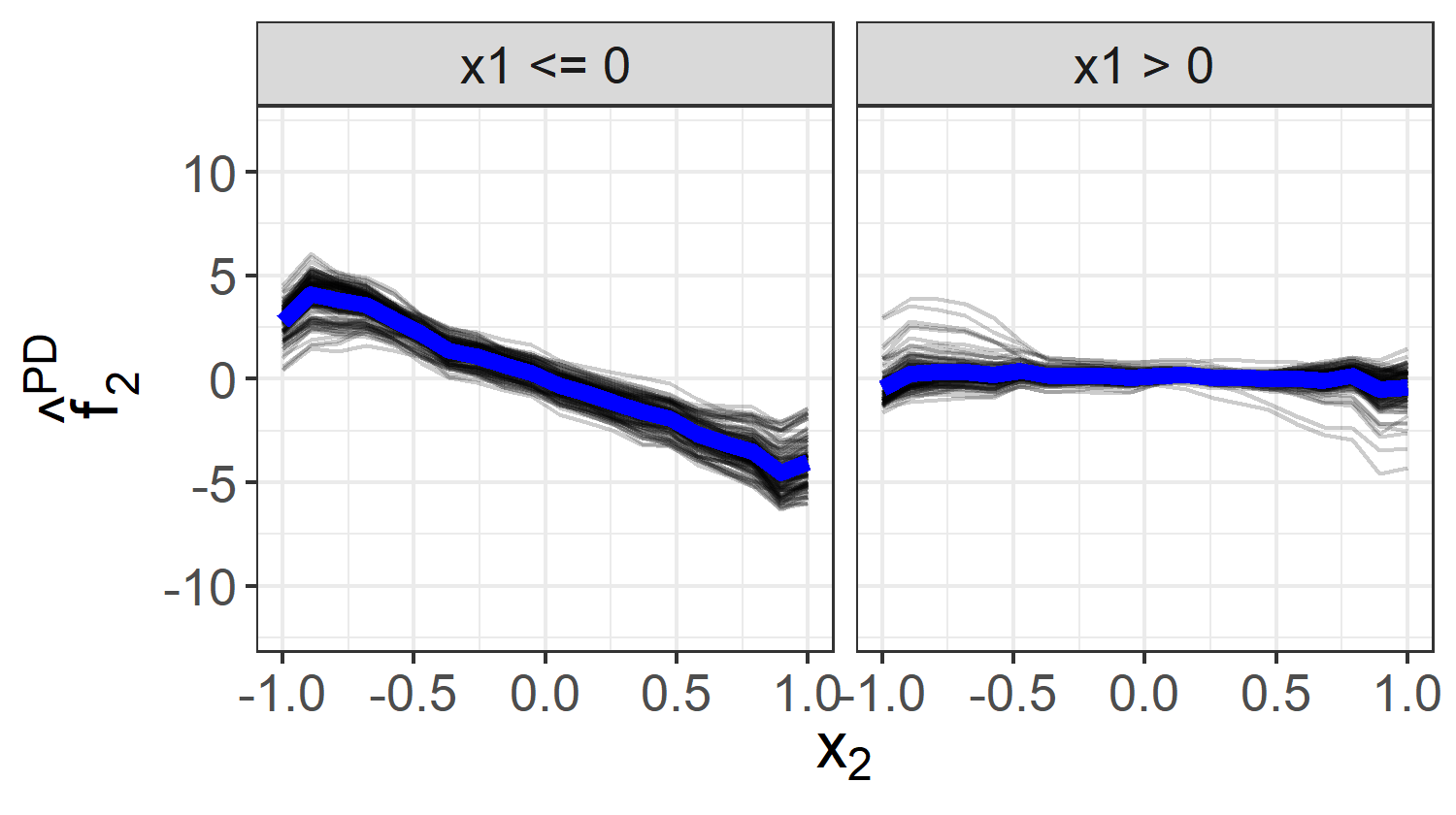}
    \vspace{.2in}
    \caption{ICE curves for $\xv_2$ grouped by REPID (black) and REPs (blue). 
    }
    \label{fig:sim1_dt}
\end{figure}
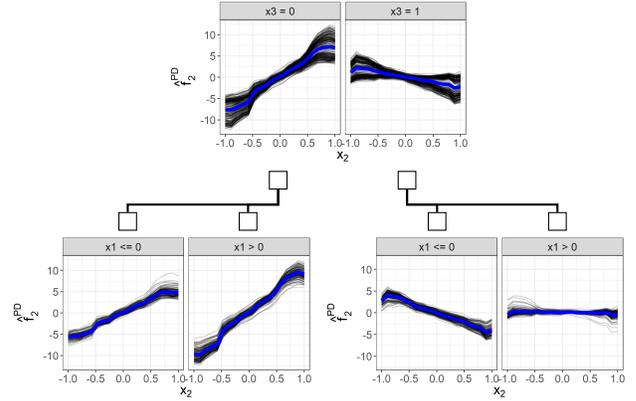

\begin{theorem}
If $\mathcal{I}(t,j) = 0$, i.e., the theoretical minimum of the split criterion is reached for a split, then the ICE curves within each of the child nodes $\mathcal{N}_l$ and $\mathcal{N}_r$ are identical to the respective REP (e.g., $\hat f^{c}(\xv_S, \xv_{C}^{(i)}) = \hat f_{S|\mathcal{N}_l}^{PD,c}(\xv_S) \quad \forall i \in \mathcal{N}_l$).
\label{theorem3}
\end{theorem}
\textit{Proof 3}
Since $\mathcal{R}_{L2}(\mathcal{N}_g) \geq 0$ $\forall g \in \{1,\ldots,G\}$, $\mathcal{I}(t,j) = 0$ implies $\hat f^{c}(\xv_S, \xv_{C}^{(i)}) = \hat f_{S|\mathcal{N}_g}^{PD,c}(\xv_S)$, $\forall i \in \mathcal{N}_g, \forall g \in \{l,r\}$.\\
\\
Applying our method to the simulation example introduced at the beginning of Section \ref{sec:rep} leads to the REPs shown in Figure \ref{fig:sim1_dt} after two splits. The first binary split divides the ICE curves of $\xv_2$ using feature $\xv_3$, which interacts most with $\xv_2$ (according to Eq.\eqref{eq:simexample}). Each of the 2 resulting regions is then split again into 2 groups by feature $\xv_1$, which also interacts with $\xv_2$. Hence, after the second split, we receive interpretable and distinct regions with REPs that represent each sub-population well.

\subsection{Quantifying Interaction Strength}
\subsubsection{Motivation}
\label{sec:quant_motiv}
Besides understanding how other features influence the marginal effect of $\xv_S$, users might be interested in how strong these interactions are and how to rank these features regarding their interaction strength with $\xv_S$.
The H-Statistic defined in Section \ref{sec:rel_work} is a global measure that quantifies the strength of interaction between two features. However, its values are influenced by the main effects of the two regarded features (see Theorem \ref{theorem4}). 
Hence, the two-way interaction with the highest H-Statistic value is not necessarily the strongest interaction, which we demonstrate in Section \ref{sec:sim_weak}. 

\begin{theorem}
The variance of the 2-dimensional mean-centered PD plot of features $\xv_j$ and $\xv_l$ ($Var(f^{PD,c}_S(\xv_S))$ with $S = \{j,l\}$) depends on the mean-centered main effects (i.e., $g^{c_S}_j(\xv_j)$ and $g^{c_S}_l(\xv_l)$) of the two features of interest $\xv_j$ and $\xv_l$. Since $Var(f^{PD,c}_S(\xv_S))$ is the denominator of the H-Statistic, which is estimated as in Eq.~\eqref{eq:hstatistic}, the H-Statistic itself also depends on the main effects of features in $S$.
The proof can be found in Appendix \ref{app:proof_t4}.
\label{theorem4}
\end{theorem}
The global interaction index proposed by \cite{greenwell2018simple} 
suffers from the same problem that we illustrate in Section \ref{sec:sim_weak} (see also Appendix \ref{app:proofs_main}).
A third method of quantifying the two-way interaction strength between features is based on SHAP interaction values (see Section \ref{sec:rel_work}).
To the best of our knowledge, SHAP interaction values have been only defined on an observational level.
Similar to the global feature importance used in \cite{lundberg2018consistent} to rank features according to their global impact in their SHAP summary plots, we suggest summarizing the individual SHAP interaction values for two features $\xv_j$ and $\xv_l$ into a global SHAP interaction index by
$$\textstyle I_{j,l}^{\text{rel}} = \frac{I_{j,l} }{\sum_{l \in \{1, \dots, p\} \setminus \{j\}} I_{j,l} } \text{ where }    I_{j,l} = \sum_{i = 1}^n |\Phi_{j,l}(\xi)|. $$
Since the absolute values $I_{j,l}$ are difficult to interpret, we prefer a relative version $I_{j,l}^{\text{rel}}$, which we call the SHAP interaction index and can be interpreted as the proportion of all two-way interactions with $\xv_j$ to which the $l$-th feature contributes.
By definition, SHAP interaction values only contain the interaction effect between $\xv_j$ and $\xv_l$. 
Hence, in contrast to the H-Statistic, varying main effects do not change the ranking of our proposed global SHAP interaction index $I_{j,l}^{\text{rel}}$.
However, both SHAP interaction indices and the H-Statistic are based on the joint distribution of the two regarded features, and hence, correlations between $\xv_S$ and features in $\xv_C$ might bias the interaction value calculated by these methods, as demonstrated in Section \ref{sec:sim_weak}. 

\subsubsection{Methodology}
\label{sec:quant_method}
Here, We derive an interaction index based on the split criterion minimized in Algorithm \ref{alg:tree} and Eq.~\eqref{eq:risk_L2}, and we prove its advantages compared to alternatives mentioned in Section \ref{sec:quant_motiv}.
Since the risk function of our split criterion is based on the variance of mean-centered ICE curves -- which measures the degree of existing feature interactions with $\xv_S$ -- we can use the achieved risk reduction after a split to quantify the interaction strength.
For better interpretability and comparability, we define the relative interaction importance for each parent node $\mathcal{N}_P$ by 
\begin{eqnarray} \textstyle
   intImp(\mathcal{N}_P) = \frac{\mathcal{R}_{L2}(\mathcal{N}_P) - (\mathcal{R}_{L2}(\mathcal{N}_l) + \mathcal{R}_{L2}(\mathcal{N}_r))}{\mathcal{R}_{L2}(\mathcal{N})} 
   \label{eq:intImp_node}
\end{eqnarray}
with $l, r \in \{1, \ldots G\}$ denoting the left and right child node of a parent node $\mathcal{N}_P$ and $\mathcal{N}$ representing the root node.
Hence, $intImp(\mathcal{N}_P)$ measures the relative risk reduction after splitting $\mathcal{N}_P$ compared to the risk within the root node $\mathcal{R}_{L2}(\mathcal{N})$.
Let $\mathcal{B}_P \subset \{1,\ldots G\}$ denote the index set of all parent nodes (i.e., all nodes that have child nodes), and let $\mathcal{B}_j \subseteq \mathcal{B}_P$ denote the subset of these parent nodes that used the regarded feature $\xv_j$ for splitting.
To obtain the relative interaction importance of feature $\xv_j$, we sum up the relative interaction importance over 
the parent nodes in $\mathcal{B}_j$:
\begin{eqnarray}
\textstyle
   intImp_j = \sum\nolimits_{P \in \mathcal{B}_j} intImp(\mathcal{N}_P).
   \label{eq:intImp_feature}
\end{eqnarray}
This principle of summing up the relative risk reduction of individual splits regarding a certain feature in order to measure the interaction strength is related to how a decision tree measures the Gini or mean decrease impurity (MDI) feature importance \citep{breiman:1984}. 
We obtain a measure that reports how important each of these features is for reducing interactions and thus obtaining more representative REPs for $\xv_S$.
Our proposed interaction importance in Eq.~\eqref{eq:intImp_feature} only depends on the interaction effects between $\xv_j$ and $\xv_S$ and not on their main effects (see Theorem \ref{theorem2}), as opposed to the H-Statistic or the interaction index of \cite{greenwell2018simple}. Furthermore, we show by Theorem \ref{theorem5} that $intImp$ -- in contrast to the H-Statistic and the SHAP interaction index $I_{j,l}^{\text{rel}}$ -- is not influenced by correlations between $\xv_S$ and $\xv_j$.
\begin{theorem}
Correlations between $X_S$ and $X_C$ do not influence the splitting procedure of REPID, since the loss function $\mathcal{L}$ of Eq.~\eqref{eq:splitting_crit_L2} does not contain a covariance term between $X_S$ and features in $X_C$. The proof can be found in Appendix \ref{app:proof_t5}.
\label{theorem5}
\end{theorem}
To determine how well the resulting REPs in the terminal nodes represent the underlying ICE curves, we derive an $R^2$ measure, which is commonly used in statistics.
The $R^2$ can be calculated by 
  $R^2 = 1 - \frac{\text{SSE(complex model)}}{\text{SSE(baseline model)}}$ 
where the baseline model is, e.g., a constant mean prediction and the $SSE$ is the sum of squared errors of the model. The measure (usually) only takes values between 0 and 1 when applied on training data. While a value of 1 indicates that the complex model fits the data perfectly, a value of 0 implies that the complex model does not outperform the baseline model.
Similar to this concept, we use the global PD plot as our baseline model.
Our complex model is the additive combination of the REPs in the terminal nodes of the final tree. Hence, each additive functional component (REP) is only valid for the specified region. The SSE of each model is measured by the variability of the underlying ICE curves.
Let $\mathcal{B}_t = \mathcal{B}_P^\complement$ denote the subset of terminal nodes in a symmetric tree.
We derive an interaction-related $R^2$ measure by aggregating the interaction importance over all parent nodes $\mathcal{B}_P$: 
\begin{eqnarray}
\textstyle
    R^2_{int} = \sum_{P \in \mathcal{B}_P} intImp(\mathcal{N}_P)
    = 1- \frac{\sum_{t \in \mathcal{B}_t} \mathcal{R}_{L2}(\mathcal{N}_t)}{\mathcal{R}_{L2}(\mathcal{N})}
\end{eqnarray}
A detailed derivation can be found in Appendix \ref{app:deriv_rsq}.

For our example, we obtain the relative interaction importance values for $\xv_2$, as stated in Table \ref{tab:intImportanceExample}. Since both child nodes after the first split use $\xv_1$ as the splitting feature, the relative interaction importance values of the two nodes can be aggregated to obtain $intImp_1 = 0.14$. It follows that REPID detects (only) the feature interactions with $\xv_2$ that have been specified in the underlying data-generating process and also ranks them in the correct order. 
The total variance after the second split is reduced by $R^2_{int} = 97.5\%$ compared to the root node, suggesting that resulting REPs are now meaningful representatives for the average marginal effect, as shown in Figure \ref{fig:sim1_dt}.
\begin{table}[thb]
\vspace{.1in}
\caption{Relative interaction importance on a node level (left) and on a feature level (right). Gray shadings indicate how $intImp_j$ is calculated from $intImp(\mathcal{N}_P)$. The parameters $d$ and $P$ indicate the tree depth and the index of the parent node, respectively.}
\vspace{.1in}
    \label{tab:intImportanceExample}
    \begin{center}
        \begin{tabular}{cc}
    \begin{tabular}{|c|c|c|c|}
    \hline
       d   & P & $\xv_j$ & $intImp(\mathcal{N}_P)$  \\\hline
       \rowcolor[gray]{.9}
        0      & 1 & $\xv_3$     & 0.835 \\
        \rowcolor[gray]{.6}
        1      & 2 & $\xv_1$     &  0.074\\
        1      & 3 & $\xv_1$     &  0.066\\\hline
    \end{tabular}\hspace*{0.5cm}
    \begin{tabular}{|c|c|}
    \hline
       $\xv_j$ & $intImp_j$  \\\hline
       \rowcolor[gray]{.9}
        $\xv_3$     & 0.835 \\
        \rowcolor[gray]{.6}
        $\xv_1$     &  0.14\\\hline
    \end{tabular}
    \end{tabular}
    \end{center}
\end{table}
\paragraph{Stop Criteria}
A possible stop criterion for the tree is to limit the maximum depth of the tree or to define a minimum number of observations for each node. 
Furthermore, we can apply a stop criterion based on the interaction importance $intImp$. Let $\mathcal{N}_g$ be the node we want to split and let $\mathcal{N}_P$ be its parent node. Then, we only split deeper if $intImp(\mathcal{N}_g) \geq \gamma \cdot intImp(\mathcal{N}_P)$, with $\gamma \in [0,1]$. In other words, we only split deeper if the improvement of the current split is at least as large as a pre-specified proportion of the improvement of the previous split. 
The suggested criteria can also be combined and the hyperparameters must be chosen by the user and usually depend on the underlying setting.

\section{SIMULATION EXAMPLES}
\label{sec:sim}

For many model-agnostic interpretation techniques -- including interaction detection methods -- ground truth information is usually not available on real-world data. Therefore, well-constructed simulation experiments with a known ground truth are often used for empirical evaluations and comparisons, while only one or few real-world datasets are used to demonstrate practical applicability 
(e.g., see \cite{friedman2008predictive}, \cite{fisher2019all}, \cite{goldstein2015peeking}, \cite{greenwell2018simple}, or \cite{aas:2021}). Hence, we follow this commonly used approach to evaluate our method using various simulation settings.

\subsection{Weaknesses of other Methods}
\label{sec:sim_weak}
\begin{figure}[htb]
    \centering
    \includegraphics[width=0.98\linewidth]{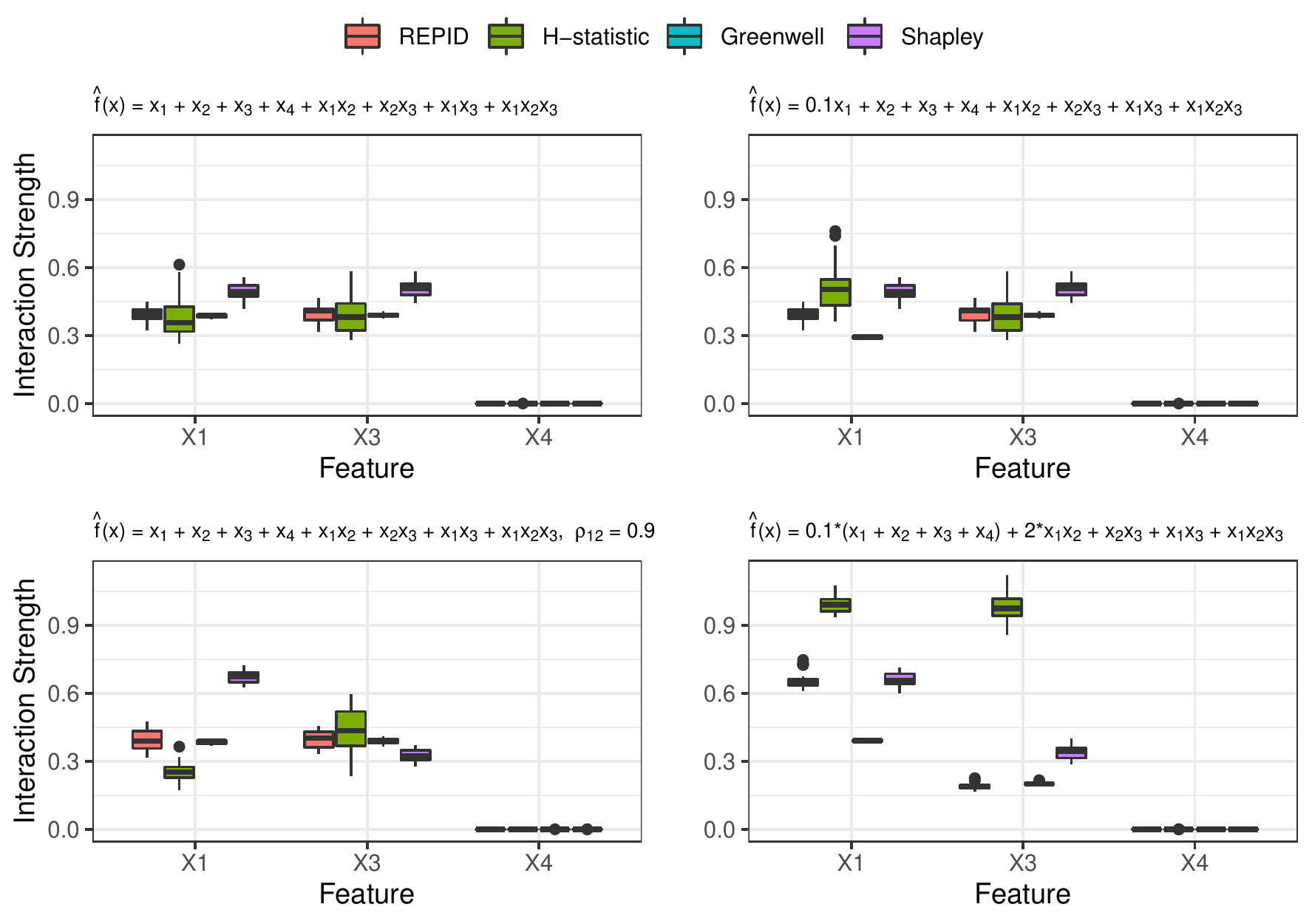}
    \vspace{.2in}
    \caption{Comparison of REPID, the H-Statistic, Greenwell's, and SHAP interaction indices for interactions between $\xv_2$ and all other features for 30 repetitions. The upper left plot shows the initial setting (1). The upper and lower right plots adjust effect sizes (2), while the bottom left plot adjusts the correlation (3).}
    \label{fig:sim_fail_hstat2}
\end{figure}

In Section \ref{sec:quant_motiv}, we described disadvantages of several interaction measures 
from a theoretical perspective.
In the following simulation example, we provide further empirical evidence.
To be able to modify the degree of the feature dependencies later on, we use a Gaussian copula to simulate the data in all settings. 
In the initial setting, we draw 1000 samples of four approximately i.i.d. random variables, which are marginally $X_1, \ldots , X_4 \sim \mathcal{U}(-1,1)$, and assume the true underlying function of $f(\xv) = r(\xv) + \epsilon$,
where $\epsilon \sim \mathcal{N}(0, (\sigma(r(\xv))\cdot 0.1)^2)$. We define the remainder by 
$
\textstyle
r(\xv) = \sum\nolimits_{j=1}^4 \xv_j + \xv_1 \xv_2 + \xv_2 \xv_3 + \xv_1 \xv_3 + \xv_1 \xv_2 \xv_3.
$
To avoid undefined interaction effects, we fit a correctly specified linear model on the data. We repeat the experiment 30 times, and each time, we measure the interaction strength between $\xv_2$ and the other three features using REPID as well as the three alternatives (the H-statistic, the Greenwell's interaction index, and the SHAP interaction index).
On three adjusted settings, we then illustrate that already small modifications of main effect sizes or feature dependencies may produce misleading results for some of the alternatives when used as a measure to rank interactions, while REPID provides correct and stable results.
For the computations, we used an equidistant grid of size 20 for REPID and Greenwell's interaction index. 
For better comparability, we used a sample size of 20 for the H-Statistic. We calculated the SHAP interaction index by aggregating the individual interaction indices for 100 randomly sampled observations, which are approximated by using 20 random permutations for all possible feature coalitions. For REPID, we combine the stop criteria described in Section \ref{sec:quant_method} as follows: We use a maximum depth of 6, a minimum number of 10 observations per node, and an improvement factor of $\gamma = 0.15$.\\
\textit{(1) Initial Setting:} The plot on the top left of Figure \ref{fig:sim_fail_hstat2} shows that, for the initial setting, all methods on average correctly assign the same interaction importance to $\xv_1$ as to $\xv_3$, 
while $\xv_4$ does not interact with $\xv_2$.\\
\textit{(2) Small main effects:}
If we reduce the main effect of $\xv_1$ to $0.1$, we observe in the top right plot of Figure \ref{fig:sim_fail_hstat2} that its interaction strength with $\xv_2$ increases on average when the H-Statistic is used. This effect can be explained by Theorem \ref{theorem4}. Hence, when main effects decrease, the proportion of the variance that explains the interaction between $\xv_1$ and $\xv_2$ increases compared to the proportion of the variance that explains the respective main effects. Also the method of Greenwell's interaction index depends on the main effect sizes. However, since Greenwell's interaction index includes the main effects in the nominator, the effect on the resulting interaction index is opposite to the one of the H-Statistic which includes the main effects in the denominator. On the other hand, the SHAP interaction index as well as REPID are only based on interaction effects, and hence, varying main effects do not change the ranking. The plot on the bottom right of Figure \ref{fig:sim_fail_hstat2} illustrates how problematic small main effects can be when the H-Statistic is applied. 
The H-Statistic leads to average interaction values close to 1 for $\xv_1$ and $\xv_3$, although the actual interaction effect of $\xv_1$ with $\xv_2$ is twice as high as that of $\xv_3$ with $\xv_2$.\\ 
\textit{(3) Dependencies between the feature of interest and other features:}
In the lower left plot of Figure \ref{fig:sim_fail_hstat2}, the correlation between $\xv_1$ and $\xv_2$ has been set to $\rho_{12} \approx 0.9$. 
Since we face a positive linear interaction effect between $\xv_1$ and $\xv_2$, a positive linear correlation between these features leads to an increasing denominator of the H-Statistic. Hence, the respective H-Statistic value decreases compared to features that are independent of $\xv_2$ (here, $\xv_3$). 
The SHAP interaction index for $\xv_1$ is higher than for $\xv_3$, since in this case, it can be shown that the interaction strength is an additive combination of the interaction effect and the covariance of the interacting features.
Conversely, Greenwell's interaction index is based on the variance of conditional marginal effects, and hence, the interaction index is not influenced by dependencies between the feature of interest and other features. The same holds for REPID, as proven with Theorem \ref{theorem5}. 

A summary of the simulation settings and key results is provided in Appendix \ref{app:sim_weak}. Detailed theoretical derivations and explanations can be found in Appendix \ref{app:proofs_weakness}.

\subsection{Comparison on More Complex Settings}
\label{sec:sim_complex}
\begin{figure}[bht]
    \centering
    \includegraphics[width=0.95\linewidth]{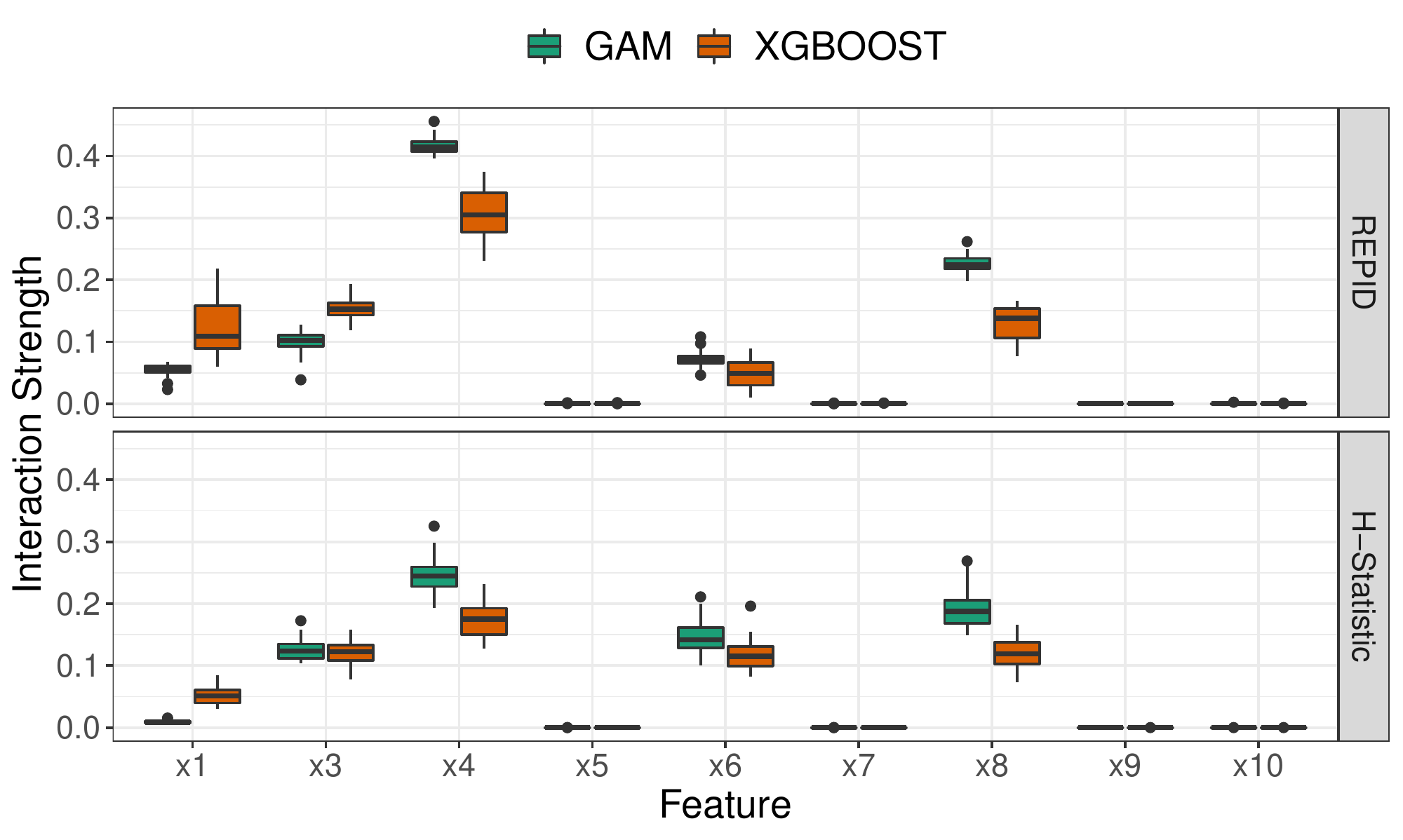}
    \vspace{.2in}
    \caption{Comparison of the interaction strength between $\xv_2$ and all other features measured by REPID (top) and the H-Statistic (bottom) on 30 repetitions. The mean (standard deviation) of the models' test performance (measured by the mean squared error) is: GAM: 0.36 (0.01), XGBOOST: 0.57 (0.11).
    }
    \label{fig:sim_nonlinear}
\end{figure}

The aim in this simulation is to show that REPID detects existing interactions correctly in a more complex non-linear setting and to compare the results to the H-Statistic. 
Analogous to \cite{hu:2020}, we draw 2000 samples of 10 independently and uniformly distributed random variables $X_1, \ldots , X_{10} \sim \mathcal{U}(-1,1)$ and assume the following true underlying function:
\begin{align}
\textstyle f(\xv) &= 6\xv_1 + \xv_2^2 - \pi^{\xv_3} + \exp^{-2\xv_4^2} + (2+|\xv_5|)^{-1} \\   
      &+ \xv_6 \log{|\xv_6|} + 2\xv_3 \mathds{1}_{(\xv_1 > 0)} \mathds{1}_{(\xv_2 > 0)} + 2\xv_2\mathds{1}_{(\xv_4 > 0)} \\ 
      &+ 4 (\xv_2\mathds{1}_{(\xv_2 > 0)})^{|\xv_6|} + |\xv_2 + \xv_8| + \epsilon
\end{align}
with $\epsilon \sim \mathcal{N}(0, 0.25)$. Hence, $\xv_2$ interacts with five other features in a more complex and non-linear way. To avoid undefined interaction effects in a fitted model, we fit a correctly specified generalized additive model (GAM) and a tree-based extreme gradient boosting model (XGBOOST) with correctly specified interaction constraints\footnote{The ``xgboost'' library \citep{Chen:2016} enables definition of which features are allowed to interact with each other.}, a learning rate of $0.1$, a maximum number of iterations of $1000$, and a maximum tree depth of $6$ on the simulated data.
The performance of each model is measured by a separately simulated test set with the same distributional assumptions of size 100,000 and is reported in Figure \ref{fig:sim_nonlinear}. We repeat the experiment 30 times, and each time, we measure the interaction strength between $\xv_2$ and the other nine features using REPID and the H-Statistic. For both methods, we again use a grid size of 20. For REPID, we apply the same stop criteria as in Section \ref{sec:sim_weak} but with a maximum tree depth of $7$ due to a more complex setting. The results are illustrated in Figure \ref{fig:sim_nonlinear}. REPID correctly identifies only the true interactions for both models. In most of the repetitions, the H-Statistic does not find an interaction between $\xv_1$ and $\xv_2$ for the GAM. A possible reason for this behavior is the rather high main effect of $\xv_1$ compared to the interaction effect (Theorem \ref{theorem4}). 
More experiments of different models and settings -- including varying values of $\lambda$ to obtain shallower or deeper trees -- can be found in Appendix \ref{app:experiments}. The experiments show that shallow trees produce fewer regions and are therefore easier to interpret. However, they might only detect the most important interactions.
Deeper trees are more likely to also identify less important interactions but are less interpretable.  

\section{REAL-WORLD EXAMPLE}
\label{sec:real_world}
\begin{figure}[tbh]
    \centering
    \includegraphics[width=0.98\linewidth]{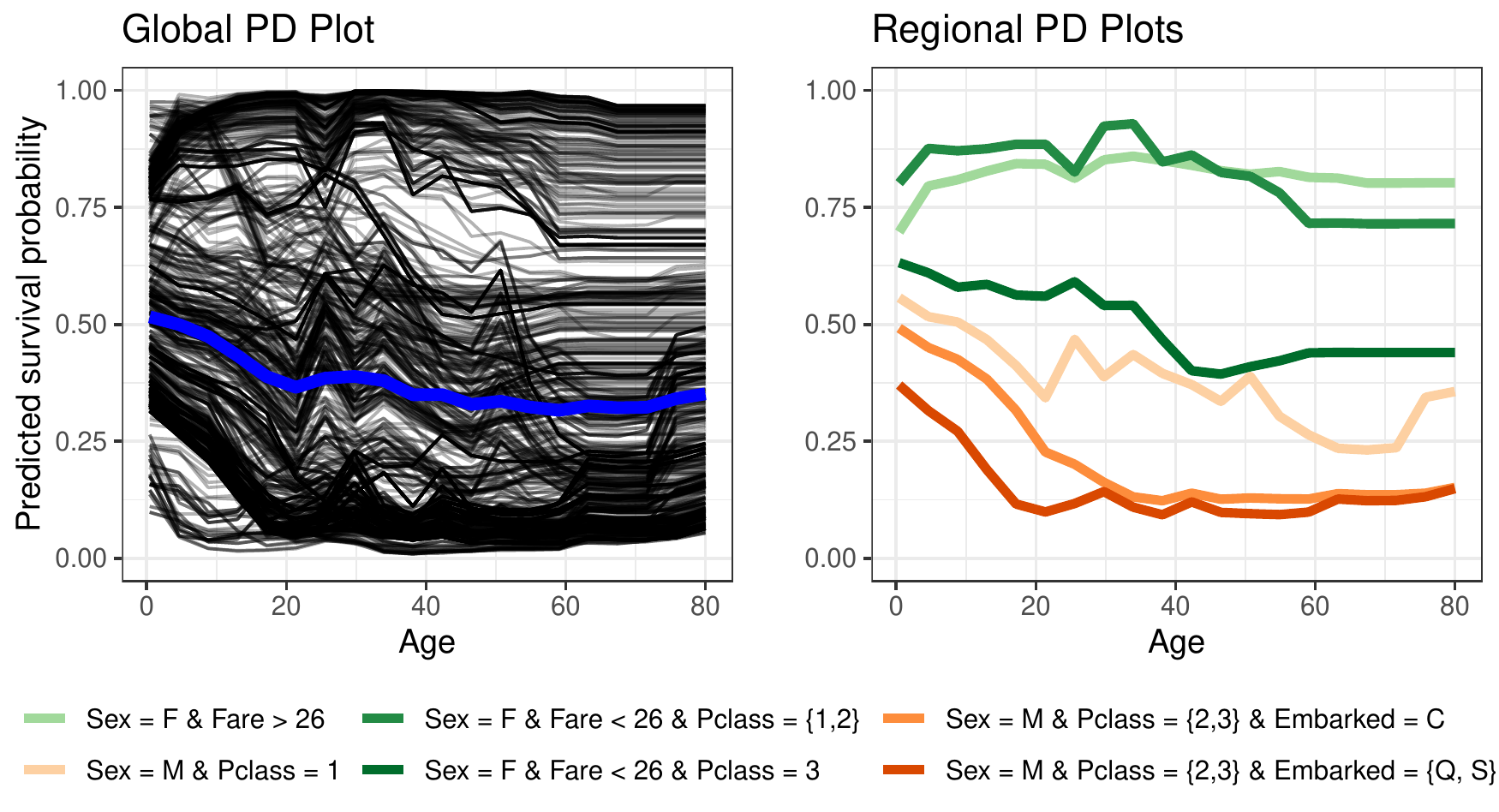}
    \vspace{.2in}
    \caption{Global PD plot (blue) including ICE curves (left) and the REPs after applying REPID (right) for the feature of interest Age of the \textit{titanic} dataset. The interaction importance $intImp_j$ between Age and the interacting features is $0.28$ (Sex), $0.17$ (Pclass), $0.13$ (Fare), $0.06$ (Embarked) and $R^2_{int} = 0.64$.
    }
    \label{fig:titanic_age}
\end{figure}

We now demonstrate the usefulness of REPID on the \textit{titanic} data \citep{dawson:1995}.
The labeled part of the dataset consists of 11 characteristics of 891 passengers of the ocean liner Titanic and a binary label if they survived. After some pre-processing steps that are described in more detail in Appendix \ref{app:titanic}, we train a RF with 500 trees on the dataset. Therefore, we obtain a balanced accuracy of $0.8$ under 5-fold cross-validation. We are interested in how the age of the passengers affects the probability of survival. The left plot in Figure \ref{fig:titanic_age} shows that, from 0 to 20 years, the PD plot for passengers continuously decreases and then flattens above 20 years. 
The ICE curves indicate that age might influence the predicted survival probability for different passengers in different ways, and thus, interactions with other features might be present. The REPs after applying REPID by using a grid size of 20, a maximum depth of 3, a minimum number of 30 observations, and $\gamma = 0.2$ are shown in the right plot of Figure \ref{fig:titanic_age}. The 3 most interacting features are Sex, Pclass (passenger class), and Fare. The green REPs show that the predicted survival probability of female passengers is on average higher compared to their male counterparts independent of their age. However, it is also visible that the probability strongly depends on the passenger's class and the fare they payed. While female passengers who payed a high fare or who belong to an upper or middle class show an overall high survival probability independent of their age (even slightly increasing until 30), the survival probability of women with a low fare and Pclass drops with age. For men from middle and lower classes, the predicted survival probability drops dramatically from 0 until 20 to 30, meaning that for the sub-population of male passengers, the chances of survival are several factors higher for children than for adults.

More real-world examples for the \textit{California housing} \citep{pace1997sparse} and the \textit{diabetes} \citep{smith1988using} datasets  are provided in Appendix \ref{app:titanic}.

\section{DISCUSSION}
We have introduced the interaction detection method REPID, which provides more representative PD plots on interpretable regions and enables quantification of feature interactions. We have proven its theoretical and empirical advantages and demonstrated how it out-performs alternatives presented in Section \ref{sec:pd_and_ice} and \ref{sec:sim}.
Unlike the H-Statistic or SHAP interaction index, REPID is not influenced by correlations between the feature of interest $\xv_S$ and other features $\xv_C$.
However, like the other methods, it might be affected if features within $\xv_C$ are correlated. Furthermore, the method might be limited if the feature of interest is, e.g., highly skewed, especially if an equidistant grid is used for computations. Possible solutions might be feature transformations or to use a sample or quantile-based grid. As our method is based on a tree-based partitioning algorithm that is known to be unstable \citep{breiman1996bagging}, the question arises whether the splitting procedure in Algorithm~\ref{alg:tree} is a potential limitation.
However, with regards to the interaction quantification, we demonstrated in Section \ref{sec:sim} that we obtain stable results when repeating the experiments multiple times. A more detailed analysis on the robustness of the method can be found in Appendix \ref{app:robustness}. 

\subsubsection*{Author Contributions (CRediT taxonomy)} 
Contributing authors: Julia Herbinger$^1$, Bernd Bischl$^2$, Giuseppe Casalicchio$^3$.
Conceptualization: 1,3; Methodology: 1,3; Project administration: 3; Formal analysis: 1,3; Writing - original draft preparation: 1; Writing - review and editing: 1,2,3; Investigation: 1,3; Visualization: 1; Validation: 1,3; Software: 1,3; Funding acquisition: 2,3; Supervision: 2,3.

\subsubsection*{Acknowledgements}
This work has been partially supported by the German Federal Ministry of Education and Research (BMBF) under Grant No. 01IS18036A, the Bavarian Ministry of Economic Affairs, Regional Development and Energy as part of the program ``Bayerischen Verbundförderprogramms (BayVFP) – Förderlinie Digitalisierung – Förderbereich Informations- und Kommunikationstechnik'' under the grant DIK-2106-0007 // DIK0260/02. The authors of this work take full responsibility for its content.

\renewcommand{\bibname}{References}
\bibliography{bib}

\clearpage
\appendix

\thispagestyle{empty}

\onecolumn \makesupplementtitle
\section{THEORETICAL EVIDENCE}
\allowdisplaybreaks
\subsection{Proofs}
\label{app:proofs_general}
Here, we provide the proofs of the Theorems defined in Section \ref{sec:pd_and_ice}. For each Theorem, we first provide a textual description in a proof sketch followed by the formal proof.\\
\textit{Note:} For our proofs, we apply the concept of functional decomposition. One concept of functional decomposition has been introduced in Section \ref{sec:rel_work}. The so-called functional ANOVA (fANOVA) decomposition is a well-known approach to decompose a function in main and interaction effects. The fANOVA decomposition defined in Section \ref{sec:rel_work} is based on \cite{hooker2004discovering}, and according to this definition, covariates must be independent to obtain a unique decomposition. However, we argue that this is not a relevant issue for our methods, since: (1) We do not try to estimate or calculate the decomposed mean-zero function terms $g_W$; we only use the (valid) assumption that a function can be decomposed as in Eq.~\eqref{eq:fANOVA} to prove our theorems. Hence, we are not directly interested in a unique solution of the decomposition. (2) Still, it is possible to relax this assumption by using the generalized fANOVA \citep{hooker2007generalized}, which is a weighted version of the ``normal'' fANOVA to address the extrapolation problem when strong correlations are present.
However, it is also possible to use another functional decomposition (e.g., as done in \cite{apley2020visualizing}) for these proofs.


\subsubsection{Proof of Theorem 1 and Corollary 1.1}
\label{app:proof_t1}

\textit{Proof Sketch}
Since $E_{X_{C}}\left [\hat f(\xv_S,\xv_C^{(i)})\right] = \hat f(\xv_S, \xv_C^{(i)})$ and if Eq.~\eqref{eq:fANOVA} holds, the fANOVA decomposition can also be applied to the $i$-th ICE curve. Since $\xv_C^{(i)}$ is constant in $i$, all fANOVA components that do not depend on $\xv_S$ can be summarized to an individual intercept shift of observation $i$ and, thus, cancelled out by mean-centering an ICE curve. The remaining term is then defined by the mean-centered main and mean-centered individual interaction effect of $\xv_S$ for observation $i$. 
Taking the expected value w.r.t. $X_C$ results in an analogous decomposition of the PD function and mean-centered PD function, respectively.

\paragraph{Proof 1}
We first derive the fANOVA decomposition of the $i$-th ICE curve $\hat f(\xv_S, \xv_C^{(i)})$ using Eq.~\eqref{eq:fANOVA} and use this decomposition to derive the mean-centered version $\hat f^c(\xv_S, \xv_C^{(i)})$ for $|S| = 1$. Therefore, we first decompose the function into main and interaction effects depending on $\xv_S$.
\textit{Note:} The term $g_0$ represents a constant intercept shift. This term is necessary to receive zero-mean functional components, i.e., e.g., $E_X[g_S(X_S)] = 0$. 


\begin{flalign}
\hat f(\xv_S, \xv_C^{(i)})
&= E_{X_C|X_C} \left[ \hat f(\xv_S, X_C)|X_C = \xv_C^{(i)}\right]&&\\ 
&= 
\underbrace{g_0}_\text{constant term} + \underbrace{g_S(\xv_S)}_\text{main effect of $\xv_S$} + 
\underbrace{\sum_{j \in C} g_j(\xv_j^{(i)})}_{\substack{\text{main effect of all other} \\ \text{features $\xv_j$ for observation $i$}}} \\ &+
\underbrace{\sum\limits_{k = 1}^{p-1}  \sum\limits_{\substack{C_k\subseteq C, \\ |C_k| = k}} g_{{C_k} \cup \{S\}}(\xv_S, \xi_{C_k})}_{\substack{\text{$(k+1)$-order interaction between} \\ \text{$\xv_S$ and $\xv_{C_k}$ for observation $i$}}}
+ \underbrace{\sum\limits_{k = 2}^{p-1}  \sum\limits_{\substack{C_k\subseteq C, \\ |C_k| = k}} g_{{C_k}}(\xi_{C_k})}_{\substack{\text{$k$-order interaction between} \\ \text{features within $C_k$ for observation $i$}}} &&
\end{flalign}


\begin{flalign}
\hat f^c(\xv_S, \xv_C^{(i)})
&= \hat f(\xv_S, \xv_C^{(i)}) - E_{X_S}\left[\hat f(X_S, \xv_C^{(i)})\right]&&\\
&= g_0 + g_S(\xv_S) + \sum_{j \in C} g_j(\xv_j^{(i)}) + \sum\limits_{k = 1}^{p-1}  \sum\limits_{\substack{C_k\subseteq C, \\ |C_k| = k}} g_{{C_k} \cup \{S\}}(\xv_S, \xi_{C_k}) + \sum\limits_{k = 2}^{p-1}  \sum\limits_{\substack{C_k\subseteq C, \\ |C_k| = k}} g_{{C_k}}(\xi_{C_k}) &&\\
&-g_0 -\underbrace{E_{X_S}\left[g_S(X_S)\right]}_\text{=0} - \sum_{j \in C} g_j(\xv_j^{(i)}) - 
 E_{X_S}\left[\sum\limits_{k = 1}^{p-1}  \sum\limits_{\substack{C_k\subseteq C, \\ |C_k| = k}} g_{{C_k} \cup \{S\}}(X_S, \xi_{C_k})\right] - \sum\limits_{k = 2}^{p-1}  \sum\limits_{\substack{C_k\subseteq C, \\ |C_k| = k}} g_{{C_k}}(\xi_{C_k})&&\\
&= \underbrace{g^{c_S}_S(\xv_S)}_{\substack{\text{mean-centered} \\ \text{main effect of $\xv_S$}}} + \underbrace{\sum\limits_{k = 1}^{p-1}  \sum\limits_{\substack{C_k\subseteq C, \\ |C_k| = k}} g_{{C_k} \cup \{S\}}(\xv_S, \xi_{C_k}) - E_{X_S}\left[\sum\limits_{k = 1}^{p-1}  \sum\limits_{\substack{C_k\subseteq C, \\ |C_k| = k}} g_{{C_k} \cup \{S\}}(X_S, \xi_{C_k})\right]}_\text{ mean-centered interaction effect of $\xv_S$ with $\xv_C^{(i)}$ for observation $i$}&&  \\
&= \underbrace{g^{c_S}_S(\xv_S)}_{\substack{\text{mean-centered} \\ \text{main effect of $\xv_S$}}} + \underbrace{\sum\limits_{k = 1}^{p-1}  \sum\limits_{\substack{C_k\subseteq C, \\ |C_k| = k}} g^{c_S}_{{C_k} \cup \{S\}}(\xv_S, \xi_{C_k})}_{\substack{\text{ mean-centered interaction effect of} \\ \text{ $\xv_S$ with $\xv_C^{(i)}$ for observation $i$}}}
\end{flalign}

\paragraph{Proof 1.1}
We first derive the fANOVA decomposition of the PD function $\hat f_S^{PD}(\xv_S)$ using Eq.~\eqref{eq:fANOVA} and use this decomposition to derive its mean-centered version $f_S^{PD,c}(\xv_S)$ for $|S| = 1$.

\begin{flalign}
f_S^{PD}(\xv_S) &= E_{X_C} \left[ \hat{f}(\xv_S, X_C)\right]&&\\ 
&= E_{X_C} \left[g_0 + g_S(\xv_S) + \sum_{j \in C} g_j(\xv_j^{(i)})  + \sum\limits_{k = 1}^{p-1}  \sum\limits_{\substack{C_k\subseteq C, \\ |C_k| = k}} g_{{C_k} \cup \{S\}}(\xv_S, X_{C_k})  + \sum\limits_{k = 2}^{p-1} \; \sum\limits_{\substack{C_k\subseteq C, \\ |C_k| = k}} g_{C_k}(X_{C_k}) \right]&&\\
&= g_0 + g_S(\xv_S) + \underbrace{E_{X_C} \left[\sum_{j \in C} g_j(\xv_j^{(i)})\right]}_{\substack{\text{expected main effect} \\ \text{of features in $\xv_C$ (=0)}}}  + E_{X_C} \left[ \sum\limits_{k = 1}^{p-1}  \sum\limits_{\substack{C_k\subseteq C, \\ |C_k| = k}} g_{{C_k} \cup \{S\}}(\xv_S, X_{C_k})\right]  + \underbrace{E_{X_C} \left[\sum\limits_{k = 2}^{p-1} \; \sum\limits_{\substack{C_k\subseteq C, \\ |C_k| = k}} g_{C_k}(X_{C_k}) \right]}_{\substack{\text{expected interaction effect} \\ \text{of features in $\xv_C$ (=0)}}}&&\\
&= g_0 +  \underbrace{g_S(\xv_S)}_\text{main effect of $\xv_S$} + \underbrace{E_{X_C} \left[ \sum\limits_{k = 1}^{p-1}  \sum\limits_{\substack{C_k\subseteq C, \\ |C_k| = k}} g_{{C_k} \cup \{S\}}(\xv_S, X_{C_k})\right]}_{\substack{\text{expected interaction effect} \\ \text{ of $\xv_S$ with $\xv_C$ w.r.t. $\xv_C$}}}
\end{flalign}

If the expected value of each decomposed term $g(\xv)$ exists and if the integral of the absolute value is finite, then Fubini's theorem can be applied, and the mean-centered PD function of $\xv_S$ for $|S| = 1$ can be derived by:
\begin{flalign}
f_S^{PD,c}(\xv_S) 
&= f_S^{PD}(\xv_S) - E_{X_S}\left[f_S^{PD}(X_S)\right]\\
&= E_{X_C} \left[ \hat f(\xv_S, X_C)\right] - E_{X_S}\left[g_0 + g_S(X_S) + E_{X_C} \left[ \sum\limits_{k = 1}^{p-1}  \sum\limits_{\substack{C_k\subseteq C, \\ |C_k| = k}} g_{{C_k} \cup \{S\}}(X_S, X_{C_k}) \right]\right]&&\\ 
&= g_0 + g_S(\xv_S) + E_{X_C} \left[ \sum\limits_{k = 1}^{p-1}  \sum\limits_{\substack{C_k\subseteq C, \\ |C_k| = k}} g_{{C_k} \cup \{S\}}(\xv_S, X_{C_k}) \right]&&\\ &- g_0 - \underbrace{E_{X_S}\left[g_S(X_S)\right]}_\text{=0} - \underbrace{E_{X_S}\left[E_{X_C} \left[\sum\limits_{k = 1}^{p-1}  \sum\limits_{\substack{C_k\subseteq C, \\ |C_k| = k}} g_{{C_k} \cup \{S\}}(X_S, X_{C_k})  \right]\right]}_{\substack{\text{expected interaction effect} \\ \text{between $\xv_S$ and $\xv_C$ (=0)}}}&&\\ 
&= \underbrace{g^{c_S}_S(\xv_S)}_{\substack{\text{mean-centered } \\ \text{main effect of $\xv_S$}}} + \underbrace{E_{X_C} \left[\sum\limits_{k = 1}^{p-1}  \sum\limits_{\substack{C_k\subseteq C, \\ |C_k| = k}} g^{c_S}_{{C_k} \cup \{S\}}(\xv_S, X_{C_k})\right]}_{\substack{\text{expected mean-centered interaction effect} \\ \text{ of $\xv_S$ with $\xv_C$ w.r.t. $\xv_C$}}}
\end{flalign}

\subsubsection{Proof of Theorem 2}
\label{app:proof_t2}


\textit{Proof Sketch}
If the function $\hat f(\xv)$ can be decomposed as in Eq.~\eqref{eq:fANOVA}, then Theorem \ref{theorem1} and Corollary \ref{corollary1} hold, and the main effect of $\xv_S$ is cancelled out when calculating $\mathcal{R}_{L2}\left(\mathcal{N}_g\right)$. The remaining term is given by the distance between the $i$-th centered interaction effect and the average centered interaction effect between $\xv_S$ and $\xv_{C}$. 

\paragraph{Proof 2}
In the risk function of Eq. \eqref{eq:risk_L2}, the squared distance between the $i$-th mean-centered ICE curve $\hat f^{c}(\xv_S, \xv_C^{(i)})$ and the respective PD function $f_S^{PD,c}(\xv_S)$ is calculated. The distance can be reduced to the following term: 
\begin{flalign}
\hat f^{c}(\xv_S, \xv_C^{(i)}) - f_S^{PD,c}(\xv_S)
&= g^{c_S}_S(\xv_S) + \sum\limits_{k = 1}^{p-1}  \sum\limits_{\substack{C_k\subseteq C, \\ |C_k| = k}} g^{c_S}_{{C_k} \cup \{S\}}(\xv_S, \xi_{C_k}) - g^{c_S}_S(\xv_S) - \sum\limits_{k = 1}^{p-1}  \sum\limits_{\substack{C_k\subseteq C, \\ |C_k| = k}} E_{X_C} \left[ g^{c_S}_{{C_k} \cup \{S\}}(\xv_S, X_{C_k})\right]&&\\
&= \sum\limits_{k = 1}^{p-1}  \sum\limits_{\substack{C_k\subseteq C, \\ |C_k| = k}} (g^{c_S}_{{C_k} \cup \{S\}}(\xv_S, \xi_{C_k}) - E_{X_C} [ g^{c_S}_{{C_k} \cup \{S\}}(\xv_S, X_{C_k})]\\
\end{flalign}


The first term is the mean-centered interaction effect of the i-th ICE curve, while the second term represents the mean-centered expected interaction effect over the joint distribution of $\xv_C$ (which is included in the mean-centered PD function, see also the decomposition of the mean-centered PD function $f_S^{PD,c}(\xv_S)$ in the proof in Appendix \ref{app:proof_t1}).
The intuition behind our split criterion is that we search for the optimal split value of a feature in $\xv_C$ that reduces the aggregated variance over all curves the most if we split according to this optimal split value. Thus, we try to find regions in the feature space $\xv_C$ where the distance between the individual centered ICE curves in this region and the respective mean-centered PD plot is as small as possible. Hence, we want to minimize the deviation of the individual interaction effect of the ICE curves in a region from the average interaction effect in the considered region. 

\subsubsection{Proof of Theorem 4}
\label{app:proof_t4}

\textit{Proof Sketch}
The two-way interaction index of the H-Statistic is calculated by dividing the variance of the difference between the centered 2-dimensional PD plot and the 1-dimensional PD plots of the two features of interest (nominator) by the variance of the centered 2-dimensional PD plot (denominator, see Eq.~\eqref{eq:hstatistic}). If Eq.~\eqref{eq:fANOVA} holds, we can apply Theorem \ref{theorem1} and Corollary \ref{corollary1}, and it can be shown that the main effects of the two features of interest are cancelled out in the nominator, but are still present in the denominator (scaling factor) of the interaction index.

\paragraph{Proof 4}
Let $S = \{j,l\}$ and $C = S^\complement$ its complement, then the 2-dimensional PD function $f^{PD}_{S}(\xv_S)$ of $\xv_j$ and $\xv_l$ is given by
\begin{flalign}
f^{PD}_{S}(\xv_S) 
&= E_{X_C} \left[ f(\xv_S, X_C)\right]&&\\ 
&=  g_0 + g_j(\xv_j) + g_l(\xv_l) + g_{jl}(\xv_{j}, \xv_{l}) + \underbrace{E_{X_{C}}  \left[\sum\limits_{k = 1}^{p-2}  \sum\limits_{\substack{C_k\subseteq C, \\ |C_k| = k}} g_{C_k}(X_{C_k})\right]}_{\substack{\text{expected interaction effect} \\ \text{of features in $\xv_C$ (=0)}}}\\
&+ E_{X_C} \left[\sum\limits_{k = 1}^{p-2}  \sum\limits_{\substack{C_k\subseteq C, \\ |C_k| = k}} g_{{C_k} \cup \{j\}}(\xv_j, X_{C_k})  + g_{{C_k} \cup \{l\}}(\xv_l, X_{C_k}) +  g_{{C_k} \cup \{S\}}(\xv_S, X_{C_k}) \right]\\
\end{flalign}

If the expected value of each decomposed term $g(\xv)$ exists, and if the integral of the absolute value is finite, then Fubini's theorem can be applied, and the mean-centred 2-dimensional PD function $f^{PD,c}_{S}(\xv_S)$ of features $\xv_j$ and $\xv_l$ can then be derived by

\begin{flalign}
f^{PD,c}_{S}(\xv_S) 
&= f^{PD}_{S}(\xv_S)  - E_{X_S} \left[f^{PD,c}_{S}(X_S) \right]&&\\
 &= g_0 +  g_j(\xv_j) + g_l(\xv_l) + g_{jl}(\xv_{j}, \xv_{l})&&\\ 
 &+E_{X_C} \left[\sum\limits_{k = 1}^{p-2}  \sum\limits_{\substack{C_k\subseteq C, \\ |C_k| = k}} g_{{C_k} \cup \{j\}}(\xv_j, X_{C_k})  + g_{{C_k} \cup \{l\}}(\xv_l, X_{C_k}) +  g_{{C_k} \cup \{S\}}(\xv_S, X_{C_k}) \right]&&\\
&- g_0 - \underbrace{E_{X_S}\left[g_j(X_j) + g_l(X_l) + g_{jl}(X_{j}, X_{l})\right]}_\text{=0} &&\\ &-\underbrace{E_{X_S}\left[E_{X_C} \left[\sum\limits_{k = 1}^{p-2}  \sum\limits_{\substack{C_k\subseteq C, \\ |C_k| = k}} g_{{C_k} \cup \{j\}}(X_j, X_{C_k})  + g_{{C_k} \cup \{l\}}(X_l, X_{C_k}) +  g_{{C_k} \cup \{S\}}(X_S, X_{C_k}) \right] \right]}_\text{expected interaction effect between $\xv_S$ and $\xv_C$ (=0)}&&\\
&= \underbrace{g^{c_S}_j(\xv_j) + g^{c_S}_l(\xv_l)}_{\substack{\text{mean-centered} \\ \text{ main effects of $\xv_S$}}} + \underbrace{g^{c_S}_{jl}(\xv_{j}, \xv_{l})}_{\substack{\text{mean-centered interaction effect} \\ \text{between $\xv_j$ and $\xv_l$}}} \\
& + \underbrace{E_{X_C} \left[\sum\limits_{k = 1}^{p-2}  \sum\limits_{\substack{C_k\subseteq C, \\ |C_k| = k}} g^{c_S}_{{C_k} \cup \{j\}}(\xv_j, X_{C_k})  + g^{c_S}_{{C_k} \cup \{l\}}(\xv_l, X_{C_k}) +  g^{c_S}_{{C_k} \cup \{S\}}(\xv_S, X_{C_k}) \right]}_{\substack{\text{expected mean-centered interaction effects} \\ \text{ between features in $\xv_S$ and features in $\xv_C$ w.r.t. $\xv_C$}}}
\end{flalign}

It follows that the H-Statistic still depends on the mean-centered main effects $g^{c_S}_j(\xv_j)$ and $g^{c_S}_l(\xv_l)$ of $\xv_j$ and $\xv_l$ in the denominator.

To calculate the nominator of the H-Statistic, we must subtract the 1-dimensional mean-centered PD functions of $x_j$ and $x_l$ as follows:

\begin{flalign}
f^{PD,c}_{S}(\xv_S) -  f^{PD,c}_{j}(\xv_j) - f^{PD,c}_{l}(\xv_l) &= 
g^{c_S}_j(\xv_j) + g^{c_S}_l(\xv_l) + g^{c_S}_{jl}(\xv_{j}, \xv_{l}) \\
& + E_{X_C} \left[\sum\limits_{k = 1}^{p-2}  \sum\limits_{\substack{C_k\subseteq C, \\ |C_k| = k}} g^{c_S}_{{C_k} \cup \{j\}}(\xv_j, X_{C_k})  + g^{c_S}_{{C_k} \cup \{l\}}(\xv_l, X_{C_k}) +  g^{c_S}_{{C_k} \cup \{S\}}(\xv_S, X_{C_k}) \right] \\
& - g^{c_S}_j(\xv_j) -  \sum\limits_{k = 1}^{p-1}  \sum\limits_{\substack{C_k\subseteq C \cup \{l\}, \\ |C_k| = k}} E_{X_{C \cup \{l\}}}\left[g^{c_S}_{{C_k} \cup \{j\}}(\xv_j, X_{C_k})\right] \\
& - g^{c_S}_l(\xv_l) -  \sum\limits_{k = 1}^{p-1}  \sum\limits_{\substack{C_k\subseteq C \cup \{j\}, \\ |C_k| = k}} E_{X_{C \cup \{j\}}}\left[g^{c_S}_{{C_k} \cup \{l\}}(\xv_l, X_{C_k})\right]
\end{flalign}

Thus, in the nominator of the H-Statistic, the variance of the calculated term is determined. This term only depends on interactions with features $\xv_j$ and $\xv_l$, while the main effects $g^{c_S}_j(\xv_j)$ and $g^{c_S}_l(\xv_l)$ that are present in the denominator are cancelled out. 

\subsubsection{Proof of Theorem 5}
\label{app:proof_t5}
\textit{Proof Sketch}
The loss function in Eq.~\eqref{eq:splitting_crit_L2}, which is used for the splitting in Algorithm \ref{alg:tree}, is calculated grid-wise. This means that we calculate the variation measured by the estimated variance (L2 loss) for each grid point $x_S^{(k)}$ with $k \in \{1,\ldots, m\}$. Hence, $\xv_S$ is not treated as a random variable but as a constant. It follows that when calculating the variance over all ICE curves on a specific grid point $x_S^{(k)}$, no covariance terms between $X_S$ and features in $X_C$ are considered.

\paragraph{Proof 5}
$\mathcal{L}(\mathcal{N}_g, x_S)$ of Eq.~\eqref{eq:splitting_crit_L2} is estimated by taking the variance over all mean-centered ICE curves within a region $\mathcal{N}_g$ for a fixed grid point of $\xv_S$.
Hence, for each grid point $k \in \{1,\ldots, m\}$, we calculate:
$$
\mathcal{L}(x_S^{(k)},\mathcal{N}_g) =
Var_{X|\mathcal{N}_g}(\hat f^c(X)|X_S = x_S^{(k)}) =
Var_{X|\mathcal{N}_g}[\hat f^c(x_S^{(k)},X_{C})]
.$$
Since $x_S^{(k)}$ is constant, it follows $Var_{X|\mathcal{N}_g}[\hat f^c(x_S^{(k)},X_{C})] = Var_{X_C|\mathcal{N}_g}[\hat f^c(x_S^{(k)},X_{C})]$, and hence, the calculated variance only depends on features in $C$ while there are no covariance terms between $X_S$ and features in $X_C$ included.

\subsection{Derivation of R Squared Measure}
\label{app:deriv_rsq}

Let $d = 0,\ldots, D$ be the depth of the tree, where $d = 0$ is the depth of the root node and $d = D$ of the leaf nodes of a symmetric tree, and $k$ defines the index of the node at each depth from left to right (starting from 0). With a slight abuse of notation, we denote $\mathcal{R}^{d}_{k}$ as the risk of the $k$-th node at depth $d$. For example, $\mathcal{R}^{0}_{0}$ is the risk of the root node ($\mathcal{R}(\mathcal{N})$).
Let $\mathcal{B}_t = \mathcal{B}_P^\complement$ denote the subset of terminal nodes in a symmetric tree.
We can derive an interaction-related $R^2$ measure by aggregating the interaction importance over all parent nodes $\mathcal{B}_P$:
\begin{flalign}
    R^2_{int} &=
    \sum_{P \in \mathcal{B_P}} intImp(\mathcal{N}_P)&&\\
    &= \frac{1}{\mathcal{R}_0^0}\cdot\sum\limits_{d=0}^{D-1}\sum\limits_{k=0}^d (\mathcal{R}^d_k - \mathcal{R}^{d+1}_{2k} - \mathcal{R}^{d+1}_{2k+1})\\
    &= \frac{1}{\mathcal{R}_0^0} \cdot (\mathcal{R}_0^0 - \sum\limits_{k=0}^{D-1}(\mathcal{R}^{D}_{2k} + \mathcal{R}^{D}_{2k+1}))\\
    &= 1- \frac{\sum\limits_{k=0}^{D-1}(\mathcal{R}^{D}_{2k} + \mathcal{R}^{D}_{2k+1})}{\mathcal{R}_0^0}\\
    &= 1- \frac{\sum_{t \in \mathcal{B}_t} \mathcal{R}(\mathcal{N}_t)}{\mathcal{R}(\mathcal{N})}
\end{flalign}

\textit{Explanation:}
According to Eq.~\eqref{eq:intImp_node}, $intImp(\mathcal{N_P})$ is defined by
$ \textstyle
   intImp(\mathcal{N}_P) = \frac{\mathcal{R}(\mathcal{N}_P) - (\mathcal{R}(\mathcal{N}_l) + \mathcal{R}(\mathcal{N}_r))}{\mathcal{R}(\mathcal{N})} 
$
which is, e.g., for the first split (using the new notation defined in this section) the same as $ \textstyle
   intImp(\mathcal{N}) = \frac{\mathcal{R}^0_0 - (\mathcal{R}^1_0 + \mathcal{R}^1_1)}{\mathcal{R}^0_0} 
$
and for the split of the first left and right child nodes (which we denote here by $\mathcal{N}_l$ and $\mathcal{N}_r$, respectively), we obtain
$ \textstyle
   intImp(\mathcal{N}_l) = \frac{\mathcal{R}^1_0 - (\mathcal{R}^2_0 + \mathcal{R}^2_1)}{\mathcal{R}^0_0} 
$
and 
$ \textstyle
   intImp(\mathcal{N}_r) = \frac{\mathcal{R}^1_1 - (\mathcal{R}^2_2 + \mathcal{R}^2_3)}{\mathcal{R}^0_0} 
$.
It follows that, after the second split ($D = 2$), $R^2_{int}$ can be calculated by
\begin{flalign}
    R^2_{int} &=
    intImp(\mathcal{N}) + intImp(\mathcal{N}_l) + intImp(\mathcal{N}_r)&&\\
    &= \frac{1}{\mathcal{R}_0^0} (\mathcal{R}^0_0 - (\mathcal{R}^1_0 + \mathcal{R}^1_1) + \mathcal{R}^1_0 - (\mathcal{R}^2_0 + \mathcal{R}^2_1) + \mathcal{R}^1_1 - (\mathcal{R}^2_2 + \mathcal{R}^2_3)\\
    &=\frac{1}{\mathcal{R}_0^0}\cdot\sum\limits_{d=0}^{1}\sum\limits_{k=0}^d (\mathcal{R}^d_k - \mathcal{R}^{d+1}_{2k} - \mathcal{R}^{d+1}_{2k+1})\\
    &= \frac{1}{\mathcal{R}_0^0} (\mathcal{R}^0_0 - (\mathcal{R}^2_0 + \mathcal{R}^2_1)) - (\mathcal{R}^2_2 + \mathcal{R}^2_3)\\
    &=\frac{1}{\mathcal{R}_0^0} \cdot (\mathcal{R}_0^0 - \sum\limits_{k=0}^{1}(\mathcal{R}^{D=2}_{2k} + \mathcal{R}^{D=2}_{2k+1}))\\
    &= 1- \frac{\sum\limits_{k=0}^{D-1}(\mathcal{R}^{D}_{2k} + \mathcal{R}^{D}_{2k+1})}{\mathcal{R}_0^0}\\
    &= 1- \frac{\sum_{t \in \mathcal{B}_t} \mathcal{R}(\mathcal{N}_t)}{\mathcal{R}(\mathcal{N})}
\end{flalign}
From the second to the fourth line of the equation, we can see that the parent nodes (besides the root node) are cancelled out when aggregating the interaction importance over all nodes. It follows that only the deviation between the root node risk and the sum over all terminal node risks remains in the nominator. The denominator is always the root node (baseline) risk.

\subsection{Explanations for Weaknesses of other Methods}
\label{app:proofs_weakness}

\subsubsection{Small Main Effects}
\label{app:proofs_main}

For REPID, we proved with Theorem \ref{theorem2} that the split criterion only depends on interaction effects with the feature of interest $\xv_S$ and is independent of main effects. 
On the other hand, according to Theorem \ref{theorem4}, the H-Statistic depends on main effects in the denominator of the H-Statistic. Since the main effect of feature $\xv_1$ is reduced from $1$ to $0.1$ in the adjusted example of Section \ref{sec:sim_weak}, the denominator of H-Statistic decreases, and hence, the overall H-Statistic value increases for feature $\xv_1$.

Since we provided proofs for REPID and for the H-Statistic, we will not go into more detail here, but instead derive explanations for the SHAP and Greenwell's interaction indices with regards to varying main effects.


\paragraph{SHAP interaction index}
By definition, SHAP interaction values only contain the interaction effect between the two features of interest and do not contain their main effects. Since we only sum up the absolute interaction values and divide them by the total amount of two-way interaction values between the feature of interest and all other features, there are also no main effects included in the global SHAP interaction index. Hence, varying main effects does not change the interaction strength / ranking calculated by the SHAP interaction index.

\textit{Example:} Due to the complexity of an increasing number of feature permutations, we show this relationship on the following simple model:
$
\hat f(\xv) = \hat\beta_1 \xv_1 + \hat\beta_2 \xv_2 + \hat\beta_{12}\xv_1 \xv_2
$
with $E(X_1) = E(X_2) = 0$.\\
In this case, we can straightforwardly calculate the individual components of the SHAP interaction value with $S = \emptyset$:
$$
f^{PD}_{S \cup \{1,2\}} (\xv_{S \cup \{1,2\}}) = \hat\beta_1 \xv_1 + \hat\beta_2 \xv_2 + \hat\beta_{12}\xv_1 \xv_2
$$
Since $E(X_1) = E(X_2) = 0$, it follows:
$$
f^{PD}_{S \cup \{1\}} (\xv_{S \cup \{1\}}) = \hat\beta_1 \xv_1 \text{   and   } f^{PD}_{S \cup \{2\}} (\xv_{S \cup \{2\}}) = \hat\beta_2 \xv_2 \text{   and   } f^{PD}_S (\xv_S) = E_X\left[\hat f(X)\right] = \hat\beta_{12} E_X\left[X_1 X_2\right]
$$

and hence, the SHAP interaction value between $\xv_1$ and $\xv_2$ is given by
\begin{flalign}
   \Phi_{1,2}(\xv) &= \frac{1}{2} (f^{PD}_{S \cup \{1,2\}} - f^{PD}_{S \cup \{1\}} (\xv_{S \cup \{1\}}) - f^{PD}_{S \cup \{2\}} (\xv_{S \cup \{2\}}) + f^{PD}_S (\xv_S))&&\\
   &= \frac{1}{2}(\hat\beta_1 x_1 + \hat\beta_2 x_2 + \hat\beta_{12}x_1 x_2 - \hat\beta_1 x_1 - \hat\beta_1 x_2 + \hat\beta_{12} E_X\left[X_1 X_2\right])\\
   &= \frac{1}{2}(\hat\beta_{12}x_1 x_2 + \hat\beta_{12} E_X\left[X_1 X_2\right])
\end{flalign}

\paragraph{Greenwell's interaction index}
\cite{greenwell2018simple} defines feature importance $i(\xv_j)$ as the standard deviation over the PD function of a feature $\xv_j$ with $m_j$ unique values as follows:
\begin{flalign}
    i(\xv_j)^2 &= \frac{1}{m_j-1} \sum\limits_{k=1}^{m_j} \left(\hat{f}^{PD}_j(x_{j}^{(k)}) - \frac{1}{m_j} \sum\limits_{k=1}^{m_j} \hat{f}^{PD}_j(x_{j}^{(k)})\right)^2&&\\
\end{flalign}
To calculate the interaction between $\xv_j$ and $\xv_l$, they define the conditional importance $i(\xv_j|\xv_l = \xv_{l}^{(i)})$ of a feature $\xv_j$ given the $t$-th unique value of $\xv_l$ as follows:
\begin{flalign}
    i(\xv_j|\xv_l &= x_{l}^{(t)})^2 = \frac{1}{m_j-1} \sum\limits_{k=1}^{m_j} \left(\hat{f}^{PD}_j(x_{j}^{(k)}|\xv_l = x_{l}^{(t)}) - \frac{1}{m_j} \sum\limits_{k=1}^{m_j} \hat{f}^{PD}_j(x_{1}^{(k)}|\xv_l = x_{l}^{(t)})\right)^2&&\\
\end{flalign}
With $m_j$ and $m_l$ being the number of unique values of $\xv_j$ and $\xv_l$, respectively,
the interaction measure $i(\xv_j,\xv_l)$ between these two features is then defined by:

\begin{flalign}
  i(\xv_j,\xv_l) &=  \frac{1}{2} \sqrt{\frac{1}{m_l-1}\sum\limits_{t=1}^{m_l} \left[ i(\xv_j|\xv_l = x_{l}^{(t)}) - \frac{1}{m_l}\sum\limits_{t=1}^{m_l} i(\xv_j|\xv_l = x_{l}^{(t)})\right]^2}&&\\
   &+ \frac{1}{2}\sqrt{\frac{1}{m_j-1}\sum\limits_{k=1}^{m_j} \left[i(\xv_l|\xv_j = x_{j}^{(k)})-\frac{1}{m_j}\sum\limits_{k=1}^{m_j} i(\xv_l|\xv_j = x_{j}^{(k)})\right]^2}
\end{flalign}


Instead of conditioning on all features in $C$ as done for ICE curves, \cite{greenwell2018simple} conditions only on the second feature of interest (e.g., $\xv_l$) to calculate the variation of PD curves for the first feature of interest (e.g., $\xv_j$). 
Hence, they first take the variation of each conditioned curve and then calculate the variation over all these curves. Since they calculate the squared distance of each conditioned PD curve to its mean, the distance still contains the main effects of the two features of interest (see Theorem \ref{theorem1}). 

\subsubsection{Dependencies between the Feature of Interest and other Features}
\label{app:proofs_featdep}
For REPID, we proved with Theorem \ref{theorem5} that the loss function of Eq.~\eqref{eq:splitting_crit_L2} (which is used for splitting) is not affected by dependencies between the feature of interest $\xv_S$ and features in $\xv_C$.

Hence, we will now derive explanations for the H-Statistic, the SHAP, and the Greenwell's interaction indices with regards to dependencies between the feature of interest and other features.


\paragraph{The H-Statistic}

The H-Statistic (which is estimated as in Eq.~\eqref{eq:hstatistic}) divides the variance of the difference between the mean-centered 2-dimensional PD plot and the two mean-centered 1-dimensional PD plots by the variance of the mean-centered 2-dimensional PD plot. Both the nominator and the denominator depend on the joint distribution of the two features of interest and, hence, also on the dependency between the two features.

\textit{Example}
Considering our simulation example in Section \ref{sec:sim_weak} with $E(X_1) = E(X_2) = E(X_3) = E(X_4) = 0$,
the mean-centered 2-dimensional PD function between $\xv_1$ and $\xv_2$ with $S = \{1,2\}$ is given by:
\begin{flalign}
 \hat{f}^{PD,c}_{S}(\xv_1,\xv_2) &=  \hat{\beta}_1 \xv_1 + \hat{\beta}_2 \xv_2 + \hat{\beta}_3 E(X_3) + \hat{\beta}_{12} \xv_1 \xv_2 + \hat{\beta}_{23} E(X_3) \xv_2 +\hat{\beta}_{13} \xv_1 E(X_3) + \hat{\beta}_{123} \xv_1 E(X_3) \xv_2&&\\ 
 &-  \hat{\beta}_1 E(X_1) - \hat{\beta}_2 E(X_2) - \hat{\beta}_3 E(X_3) - \hat{\beta}_{12} E_{X_S}\left[X_1 X_2\right] - \hat{\beta}_{23} E(X_3) E(X_2) -\hat{\beta}_{13} E(X_1) E(X_3)\\
 &- \hat{\beta}_{123} E_{X_S}\left[X_1 X_2\right] E(X_3)  \\
 &= \hat{\beta}_1 \xv_1 + \hat{\beta}_2 \xv_2+ \hat{\beta}_{12} (\xv_1 \xv_2 - E_{X_S}\left[X_1 X_2\right])
\end{flalign}

Calculating the denominator by taking the variance
\begin{flalign}
 Var(\hat{f}^{PD,c}_{S}(\xv_1,\xv_2)) &=  E\left[(\hat{\beta}_1 X_1 + \hat{\beta}_2 X_2+ \hat{\beta}_{12} (X_1 X_2 - E_{X_{S}}\left[X_1 X_2\right]))^2\right] &&\\
 &- E\left[\hat{\beta}_1 X_1 + \hat{\beta}_2 X_2+ \hat{\beta}_{12} (X_1 X_2 - E_{X_{S}}\left[X_1 X_2\right])\right]^2\\ 
 &= E\left[\hat{\beta}_1^2 X_1^2 + 2\hat{\beta}_1 \hat{\beta}_2 X_1 X_2 + \hat{\beta}_2^2 X_2^2 + 2\hat{\beta}_1 \hat{\beta}_{12} X_1^2 X_2 + 2\hat{\beta}_2 \hat{\beta}_{12} X_1 X_2^2\right] \\
 & +E\left[-2 \hat{\beta}_1 \hat{\beta}_{12} X_1 E_{X_{S}}\left[X_1 X_2\right]  - 2 \hat{\beta}_2 \hat{\beta}_{12} X_2 E_{X_{S}}\left[X_1 X_2\right] + \hat{\beta}_{12}^2 X_1^2 X_2^2\right] \\
 & + E\left[-2 \hat{\beta}_{12}^2 X_1 X_2 E_{X_{S}}\left[X_1 X_2\right] + \hat{\beta}_{12}^2  E_{X_{S}}\left[X_1 X_2\right]^2\right]\\
 &= \hat{\beta}_1^2 Var(X_1) + \hat{\beta}_2^2 Var(X_2)+ \hat{\beta}_{12}^2Var(X_1X_2)\\
 &+ 2 \hat{\beta}_1 \hat{\beta}_2 Cov(X_1,X_2) + 2 \hat\beta_1\hat\beta_{12} Cov(X_1^2, X_2)  + 2 \hat\beta_2\hat\beta_{12} Cov(X_1, X_2^2)
\end{flalign}

in the nominator, we subtract the mean-centered 1-dimensional PD functions (i.e., $\hat{f}^{PD,c}_{1}(\xv_1) = \hat \beta_1 \xv_1$ and $\hat{f}^{PD,c}_{2}(\xv_2) = \hat \beta_2 \xv_2$) and take the variance, which results in
\begin{flalign}
 &E_X\left[ \hat{\beta}_{12}^2 (X_1 X_2 - E_{X_{S}}\left[X_1 X_2\right]))^2\right] 
 - E_X\left[\hat{\beta}_{12} (X_1 X_2 - E_{X_{S}}\left[X_1 X_2\right])\right]^2 \\
 &= E_X\left[ \hat{\beta}_{12}^2 X_1^2 X_2^2
 -2 \hat{\beta}_{12}^2 X_1 X_2 E_{X_{S}}\left[X_1 X_2\right] + \hat{\beta}_{12}^2  E_{X_{S}}\left[X_1 X_2\right]^2\right]&&\\
 &= \hat{\beta}_{12}^2Var(X_1X_2)\\
 &= \hat{\beta}_{12}^2 (Var(X_1)V(X_2)) - Cov(X_1,X_2)^2 + Cov(X_1^2,X_2^2) )
\end{flalign}

It follows that by increasing the correlation between $\xv_1$ and $\xv_2$ to $\rho_{12} = 0.9$, the denominator of the H-Statistic increases compared to the nominator for the given example, and hence, the H-Statistic value between $\xv_1$ and $\xv_2$ decreases compared to the H-Statistic value between $\xv_2$ and $\xv_3$.

Some general rules that were applied here:
\begin{itemize}
    \item [1] Rearrangement of variance formula for functions: $Var(g(X)) = E\left[g(X)^2\right] - (E\left[g(X)\right]^2$
    \item [2] Expected value of a product of two random variables: $E\left[X_1 X_2\right]  =  E\left[X_1\right] E\left[X_2\right] + Cov(X_1,X_2)$ which reduces for $E(X_1) = E(X_2) = 0$ to $E\left[X_1 X_2\right]  =   Cov(X_1,X_2)$
    \item [3] Variance of a product of two random variables: $V(XY) = E\left[X^2Y^2\right] - (E\left[XY\right]^2 = Cov(X^2,Y^2) + (V(X) + (E\left[X\right]^2)(V(Y) + (E\left[Y\right]^2) - (Cov(X,Y) + E\left[X\right]E\left[Y\right])^2$ which reduces for $E\left[X\right] =E\left[Y\right] = 0$ to $V(XY) = Cov(X^2,Y^2) + V(X) V(Y) - Cov(X,Y)^2$
\end{itemize}

\paragraph{SHAP interaction index}
SHAP interaction values -- and with that, also the (global) SHAP Interaction index -- depend on the correlation between the two features of interest, since we consider the joint distribution of the features as we do for the H-Statistic.

\textit{Example}
In Appendix \ref{app:proofs_main}, we derived the SHAP interaction value for a simple linear model of two features with a positive linear interaction between these features, which resulted in
\begin{eqnarray*}
   \Phi_{1,2}(\xv) &= 
   \frac{1}{2} (\hat\beta_{12}\xv_1 \xv_2 + \hat\beta_{12} E_X\left[X_1 X_2\right])
\end{eqnarray*}
Hence, if $\xv_1$ and $\xv_2$ are positively correlated as in our example in Section \ref{sec:sim_weak}, then $E_X\left[X_1 X_2\right] > 0$, while this term is 0 if the two features are independent. This is why $\xv_1$ shows a higher interaction value than $\xv_3$ in the referred simulation study.

\paragraph{Greenwell's interaction index}
Similarly to our approach, the Greenwell's interaction index conditions on one of the two features of interest. They calculate the variance w.r.t. the other feature of interest, and vice versa. Hence, the dependency between the two regarded features does not influence the resulting interaction index.

\section{EMPIRICAL EVIDENCE}
\label{app:empirical}

In this section, we provide more empirical evidence for the usefulness of REPID. 
We will further analyze the nonlinear simulation setting described in Section \ref{sec:sim_complex} and will also look at a linear example where interactions can clearly be ranked. Furthermore, we analyze the influence of the improvement parameter $\gamma$ used as stop criterion and provide some evidence for the robustness of our method in Section \ref{app:robustness}. In Section \ref{app:titanic}, we clarify the pre-processing steps of the real-world example that was analyzed in Section \ref{sec:real_world}.

\paragraph{Infrastructure}

All experiments only require CPUs (and no GPUs) and were computed on a Linux cluster (see Table~\ref{tab:cluster}). 

\begin{table}[ht]
\vspace{.1in}
\caption{Description of the infrastructure used for the experiments in this paper. }
\label{tab:cluster}
\vspace{.2in}
\centering
    \begin{tabular}{ll}
        \toprule
         \multicolumn{2}{c}{Computing Infrastructure} \\ \midrule
         Type & Linux CPU Cluster \\ 
         Architecture & 28-way Haswell-EP nodes \\
         Cores per Node & 1 \\
         Memory limit (per core) & 2.2 GB \\ \bottomrule
    \end{tabular}
\end{table}


\subsection{Overview on Weaknesses of other Methods}
\label{app:sim_weak}
In Table \ref{tab:simSummary}, we provide a brief overview of the simulation setting, including a sensitivity analysis that we performed in Section \ref{sec:sim_weak}. The table shows that only REPID provides on average correct ranks for all settings, while the other state-of-the-art methods provide for at least one of the settings a wrong ranking (on average).

\begin{table}[thb]
\vspace{.1in}
\caption{Summary table of settings and key results of the simulation study in Section \ref{sec:sim_weak}. The column ``Setting'' refers to the setting number in Section \ref{sec:sim_weak}. The second column refers to the adjustments made in the setting compared to the initial setting. The other four columns show if the average ranks (r) of the feature interactions with the feature of interest ($\xv_2$) are correct (meaning that the ranks are the same as the ranks of the underlying data-generating process and fitted linear model) or if they are wrong (different from the ranks in the data-generating process and fitted linear model).} 
\vspace{.1in}
    \label{tab:simSummary}
    \begin{center}
       
    \begin{tabular}{|p{1cm}|p{4cm}|p{2.2cm}|p{2.2cm}|p{2.2cm}|p{2.2cm}|}
    \hline
       Setting   & Adjustment & REPID & H-Statistic & Greenwell & Shapley  \\\hline
        (2)      & $\beta_1 = 0.1$ (initial: 1) & \textbf{correct} $r(x_1) = r(x_3)$     & \textbf{wrong} $r(x_1) > r(x_3)$ & \textbf{wrong} $r(x_1) < r(x_3)$ & \textbf{correct}\\
        (2) & $\beta_1 = \beta_2 = \beta_3 = \beta_4 = 0.1$ and $\beta_{12} = 2$ (initial: 1) & \textbf{correct} $r(x_1) > r(x_3)$ & \textbf{wrong} $r(x_1) = r(x_3)$ & \textbf{correct} & \textbf{correct}\\
        (3) & $\rho_{12}$ = 0.9 (initial: 0) & \textbf{correct} $r(x_1) = r(x_3)$ & \textbf{wrong} $r(x_1) < r(x_3)$ & \textbf{correct} & \textbf{wrong} $r(x_1) > r(x_3)$\\
  \hline
    \end{tabular}
 
    \end{center}
\end{table}
\subsection{Further experiments}
\label{app:experiments}

\begin{figure}[bht]
    \centering
    \includegraphics[width=0.48\linewidth]{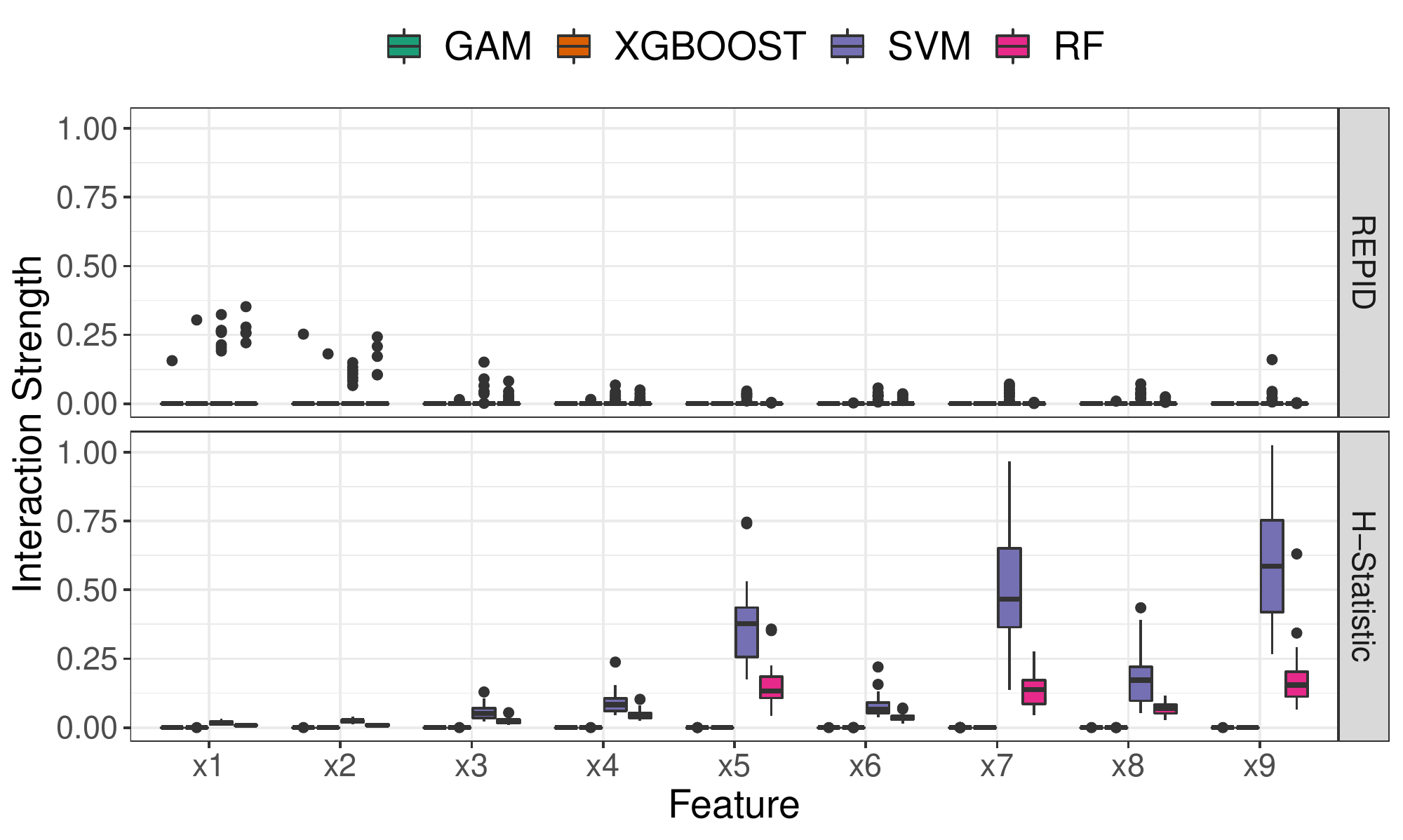}
     \includegraphics[width=0.48\linewidth]{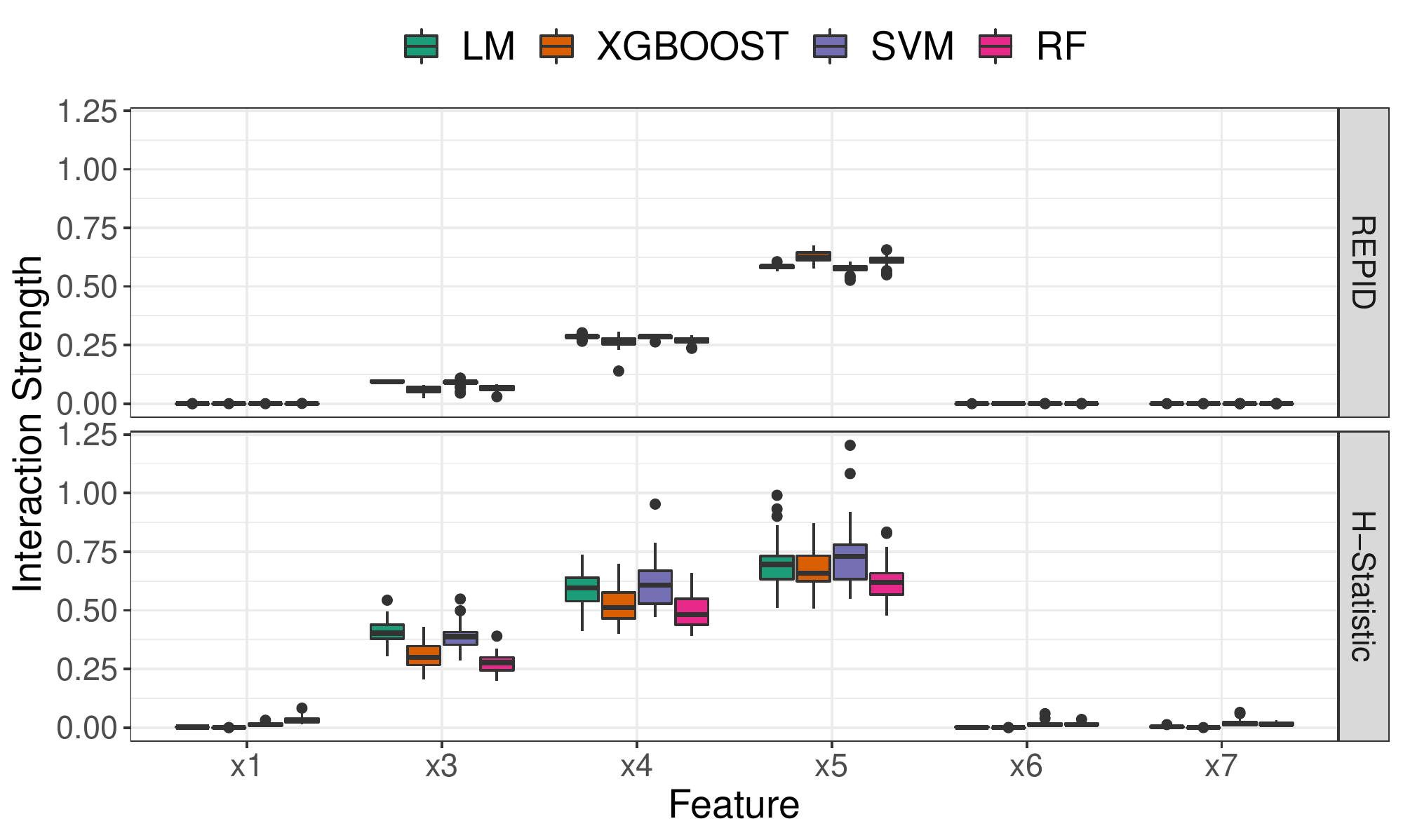}
    \vspace{.2in}
    \caption{Left (right): The figure compares the interaction strength between $\xv_{10}$ ($\xv_2$) and all other features measured by REPID and the H-Statistic for 4 different models on 30 repetitions of the described nonlinear (linear) simulation setting.}
    \label{fig:app_complex1}
\end{figure}

\paragraph{Nonlinear example}
In Section \ref{sec:sim_complex}, we compared REPID and the H-Statistic for the interactions between the most interacting feature $\xv_2$ and the other nine features of the simulation setting described in the referred section. In addition to the correctly specified GAM and XGBOOST model from Section \ref{sec:sim_complex}, we now also compare the results to two other ML models: an RF with 500 trees -- the mean and standard deviation of the models' test performance (measured by the mean squared error) is 1.01 and 0.16 -- and a support vector machine (SVM) using 
epsilon support vector regression with a Gauss kernel, $C = 1$ and $\epsilon = 0.1$ -- the mean and standard deviation of the models' test performance (measured by the mean squared error) is 0.76 and 0.07. The left plot in Figure \ref{fig:app_complex1} shows the same illustration as in Figure \ref{fig:sim_nonlinear} for the interactions between the non-influential feature $\xv_{10}$ and all other features. For the correctly specified GAM and XGBOOST model, both methods do -- as expected -- on average not find any interactions. While REPID on average also recognizes that there are no interactions present between $\xv_{10}$ and all other features for the SVM and RF models, the H-Statistic finds some higher interactions, especially for the SVM. A possible explanation is that $\xv_{10}$ does also not influence the target by a main effect in the underlying function, and hence, possible small found interaction effects might lead to high H-Statistic values. The outliers for some features when REPID is applied are possibly because the total variation of mean-centered ICE curves for non-influential features are rather small, and hence, relative loss reduction values might be high, although the absolute values are small. A potential solution to prevent these outliers is to extend the stop criterion by, e.g., a minimum absolute loss reduction constraint.

\begin{figure}[bht]
    \centering
    \includegraphics[width=0.48\linewidth]{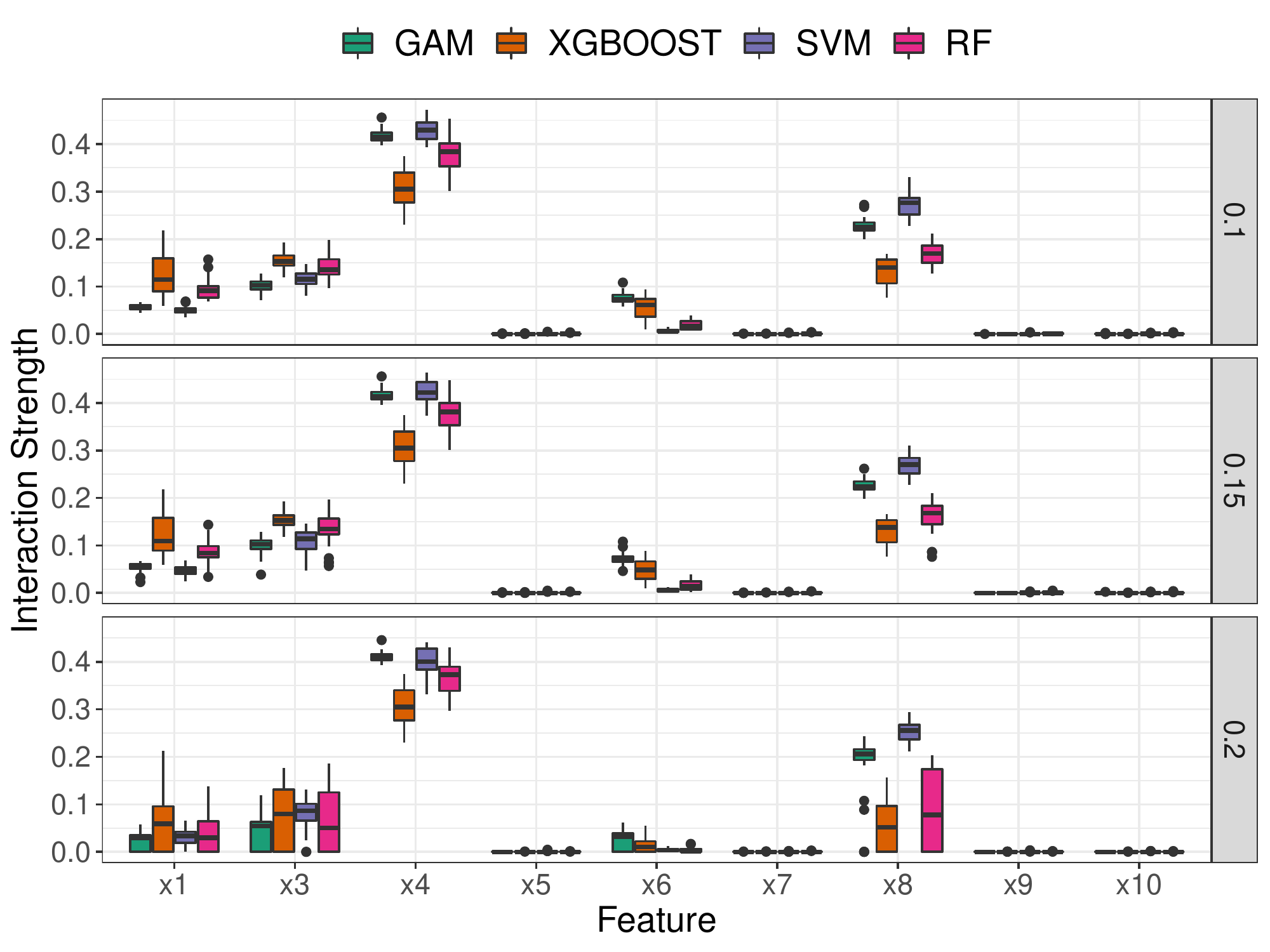}
     \includegraphics[width=0.48\linewidth]{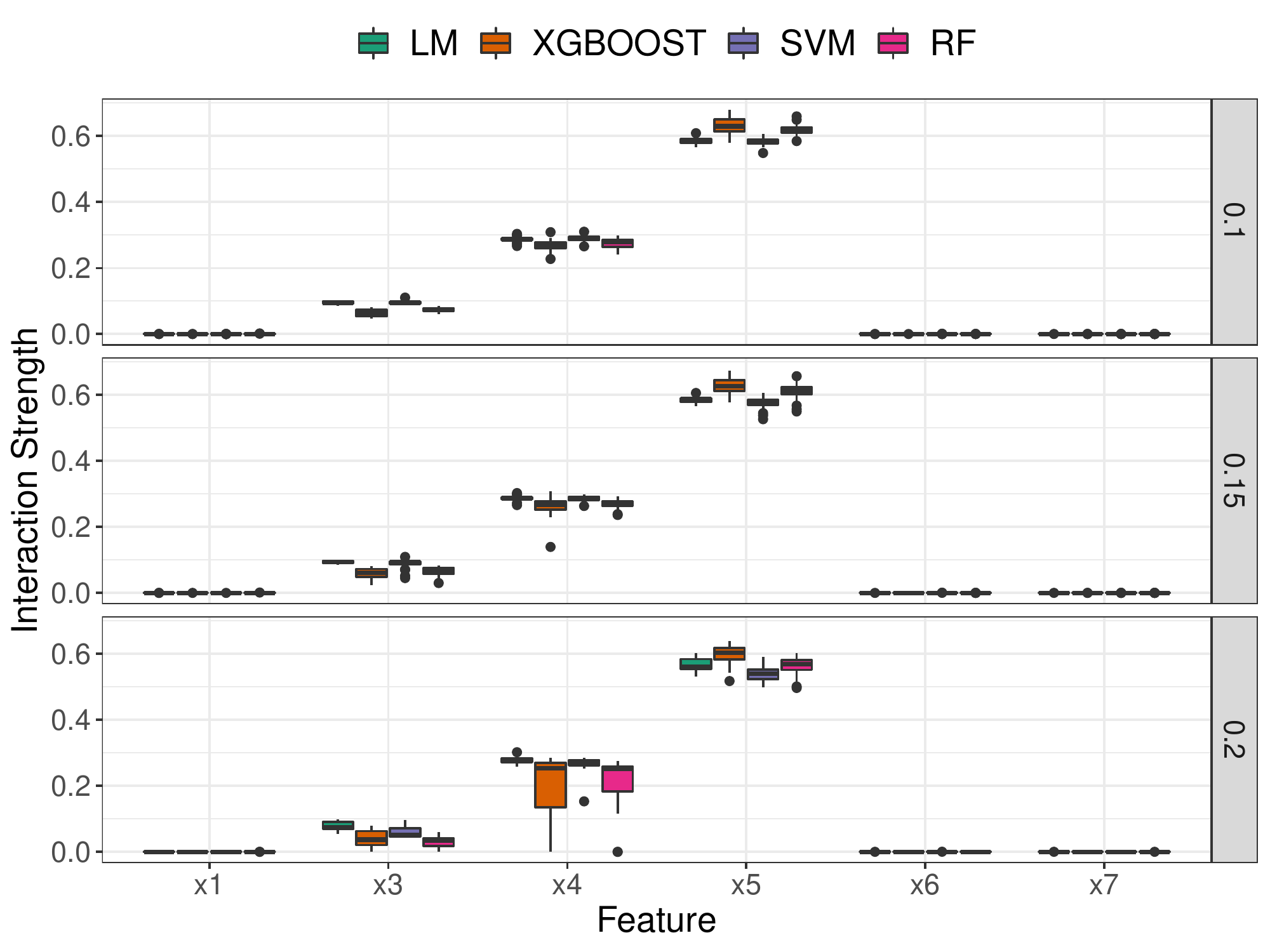}
    \vspace{.2in}
    \caption{Left (right): The figure compares the interaction strength between $\xv_2$ and all other features measured by REPID for 3 different improvement parameter values: $\gamma = 0.1$ (top), $\gamma = 0.15$ (middle), and $\gamma = 0.2$ (bottom) for 4 different models on 30 repetitions of the described nonlinear (linear) simulation setting.}
    \label{fig:app_complex2}
\end{figure}

In the left plot in Figure \ref{fig:app_complex2}, we analyzed the influence of the improvement parameter $\gamma$ on the interaction strength. The difference between the threshold $\gamma = 0.15$, which we chose in Section \ref{sec:sim_complex}, and $\gamma = 0.1$ is rather small, while it becomes more difficult to detect the smaller interactions with $\gamma = 0.2$. The smaller we choose $\gamma$ to be, the deeper we split, and the less interaction variance remains in the final terminal nodes. Therefore, the obtained interaction strengths are more precise, and hence, our results seem to be more robust for different repetitions\footnote{The more robust results are shown by smaller interquartile ranges of boxplots in Figure \ref{fig:app_complex2}.}. However, the deeper we split, the more final regions we obtain, which makes it more difficult to visually analyze the influence of the interactions on the marginal effect of the feature of interest. Hence, how to set the improvement parameter $\gamma$ depends on the question the user would like to answer.

\paragraph{Linear example}
We now look at a further simulation example with only linear interaction effects between numeric features, which makes it possible to clearly rank the interactions between the feature of interest and all other features. Therefore, we draw 2000 samples of seven independent random variables, which are distributed as follows: $X_1,\ldots, X_5 \sim \mathcal{U}(-1,1)$, $X_6 \sim \mathcal{N}(0,4)$ and $X_7 \sim \mathcal{N}(2,9)$. The true underlying relationship is defined by $f(\xv) = r(\xv)+\epsilon$, where the remainder $r(\xv)$ is given by
\begin{equation}
    r(\xv) = \xv_1 + 4\xv_2 + 3\xv_2 \xv_3 + 5\xv_2 \xv_4 + 7\xv_2 \xv_5
\end{equation}
and $\epsilon \sim \mathcal{N}(0, (\sigma(r(\xv))\cdot 0.1)^2)$.
Hence, $\xv_5$ interacts most with $\xv_2$, followed by $\xv_4$ and then $\xv_3$. We fitted a linear model (LM) and an XGBOOST model with interaction constraints as well as an SVM and RF using the same configurations as for the nonlinear example on the simulated data. We repeated the experiment 30 times to quantify the interaction strength between $\xv_2$ and all other features using REPID and the H-Statistic.\footnote{The mean (standard deviation) of the models' test performance (measured by the mean squared error) is for the LM: 0.15 (0.002), XGBOOST: 0.6 (0.22), SVM: 0.31 (0.069) and RF: 1.43 (0.34).} We use the same specifications for the models' and interaction detection methods' hyperparameters as used in Section \ref{sec:sim_complex}. The right plot in Figure \ref{fig:app_complex1} illustrates that both methods on average find the correct ranking of the feature interactions. However, REPID shows almost no variation over all repetitions and hence leads to more stable and clearer ranking results than the H-Statistic.
In the right plot of Figure \ref{fig:app_complex2}, the impact of the improvement parameter $\gamma$ is shown. However, for this example, we barely see a difference between the different choices of $\gamma$, which might be due to the simplicity of the setting and hence that no deep trees are necessary to receive stable results for the interaction strength.

\subsection{Robustness analysis}
\label{app:robustness}
An oft-stated limitation of the usage of decision trees is that they do not provide robust results. 
In Section \ref{sec:sim} and \ref{app:experiments}, we already showed that REPID provides robust results with regards to quantifying the interaction strength for different simulation settings.
To investigate the robustness itself of the fitted trees, we extract and analyze the splits of the first three levels (depths) of the tree for the nonlinear example of Section \ref{sec:sim_complex}, which is the most complex analyzed example of all examples in this paper. 
The frequencies of the features used at each split for the 30 repetitions is shown in Table \ref{tab:robustness} for each of the fitted models.
For all repetitions and for all models, $\xv_4$ was always chosen as the first splitting feature, with an average split value very close to 0, which shows only small variations (sd values). Furthermore, all models chose most often $\xv_8$ for all nodes in the second level and $\xv_3$ for all nodes in the third level of the tree. For the GAM that was correctly specified according to the true underlying function, the splits for the first three levels of the fitted decision tree barely differ. On the other hand, the SVM and the RF show higher variations. However, these models might have learned different interaction effects for different repetitions, and hence, it might be reasonable to receive different splits and REPs. The XGBOOST model also varies more than the GAM, which might be due to the fact that the GAM has a better and less variable model performance compared to the XBGOOST model, and hence, effect sizes might also vary less (see Figure~\ref{fig:sim_nonlinear}). 
However, for all models, the feature chosen most often in each node is the same. It follows that REPID seems to provide robust results with regards to the interaction strength and the upper levels of the fitted tree if the same interactions have been learned by the ML models we want to explain.
\begin{table}[ht]
\caption{Summary of the split information of the first three levels (depths) of the trees fitted by applying REPID to the simulation example stated in Section \ref{sec:sim_complex} for the 30 repetitions of the 4 models (GAM, XGBOOST, SVM, RF). The column ``Depth'' indicates the tree depth while the column ``Node ID'' indicates the respective node of this depth from left to right. The columns ``Feature'' and ``Share'' provide information of how often which feature was chosen for splitting in the respective node. The last two columns contain the mean and standard deviation of the respective split value. The coloring indicates the feature that was chosen most often for each node, where the different colors belong to the different tree depths.}
\centering
\vspace{.1in}

\begin{footnotesize}
\begin{tabular}{lrrrrrr}
  \hline
Model & Depth & Node ID & Feature & Share & Split value mean & Split value sd \\ 
  \hline
  \multirow{10}{4em}{GAM} & 1 &   1 & \cellcolor[HTML]{AAACED}x4 & 1.00 & 0.01 & 0.02 \\ 
 & 2 &   1 & \cellcolor{cyan} x8 & 1.00 & 0.02 & 0.10 \\ 
 & 2 &   2 & \cellcolor{cyan}x8 & 1.00 & -0.00 & 0.10 \\ 
 & 3 &   1 & \cellcolor{pink}x3 & 1.00 & -0.04 & 0.14 \\ 
 & 3 &   2 & x8 & 0.03 & 0.31 &  \\ 
 & 3 &   2 & \cellcolor{pink}x3 & 0.97 & 0.03 & 0.16 \\ 
 & 3 &   3 & x8 & 0.03 & -0.37 &  \\ 
 & 3 &   3 & \cellcolor{pink}x3 & 0.97 & -0.05 & 0.16 \\ 
 & 3 &   4 & x8 & 0.07 & 0.38 & 0.07 \\ 
 & 3 &   4 & \cellcolor{pink}x3 & 0.93 & -0.01 & 0.14 \\ \hline
 \multirow{13}{4em}{XGBOOST} &  1 &   1 & \cellcolor[HTML]{AAACED}x4 & 1.00 & 0.00 & 0.01 \\
 & 2 &   1 &  \cellcolor{cyan}x8 & 0.63 & -0.06 & 0.11 \\ 
 & 2 &   1 & x3 & 0.37 & -0.02 & 0.25 \\ 
 & 2 &   2 &  \cellcolor{cyan}x8 & 0.77 & -0.05 & 0.11 \\ 
 & 2 &   2 & x3 & 0.23 & 0.01 & 0.11 \\ 
 & 3 &   1 & x8 & 0.17 & -0.06 & 0.16 \\ 
 & 3 &   1 & \cellcolor{pink}x3 & 0.63 & 0.03 & 0.14 \\ 
 & 3 &   1 & x1 & 0.20 & -0.01 & 0.10 \\ 
 & 3 &   2 & x8 & 0.07 & -0.19 & 0.01 \\ 
 & 3 &   2 & \cellcolor{pink}x3 & 0.63 & 0.03 & 0.21 \\ 
 & 3 &   2 & x1 & 0.30 & 0.01 & 0.06 \\ 
 & 3 &   3 & x8 & 0.10 & 0.13 & 0.22 \\ 
 & 3 &   3 & \cellcolor{pink}x3 & 0.77 & -0.04 & 0.19 \\ 
 & 3 &   3 & x1 & 0.13 & -0.02 & 0.06 \\ 
 & 3 &   4 & x8 & 0.03 & 0.00 &  \\ 
 & 3 &   4 & \cellcolor{pink}x3 & 0.77 & 0.08 & 0.20 \\ 
 & 3 &   4 & x1 & 0.20 & 0.06 & 0.07 \\ \hline
\multirow{13}{4em}{SVM} &  1 &   1 & \cellcolor[HTML]{AAACED}x4 & 1.00 & -0.03 & 0.07 \\ 
 & 2 &   1 & \cellcolor{cyan}x8 & 1.00 & -0.02 & 0.10 \\ 
 & 2 &   2 & \cellcolor{cyan}x8 & 1.00 & -0.11 & 0.10 \\ 
 & 3 &   1 & x4 & 0.23 & -0.45 & 0.06 \\ 
 & 3 &   1 & x8 & 0.03 & -0.55 &  \\ 
 & 3 &   1 & \cellcolor{pink}x3 & 0.73 & -0.02 & 0.15 \\ 
 & 3 &   2 & x4 & 0.37 & -0.50 & 0.09 \\ 
 & 3 &   2 & \cellcolor{pink}x3 & 0.63 & -0.15 & 0.13 \\ 
 & 3 &   3 & x4 & 0.20 & 0.35 & 0.07 \\ 
 & 3 &   3 & x8 & 0.07 & -0.61 & 0.00 \\ 
 & 3 &   3 & \cellcolor{pink}x3 & 0.73 & -0.18 & 0.16 \\ 
 & 3 &   4 & x4 & 0.07 & 0.24 & 0.05 \\ 
 & 3 &   4 & \cellcolor{pink}x3 & 0.93 & -0.21 & 0.15 \\ \hline
\multirow{12}{4em}{RF} &
  1 &   1 & \cellcolor[HTML]{AAACED}x4 & 1.00 & 0.00 & 0.02 \\ 
 & 2 &   1 & \cellcolor{cyan}x8 & 0.70 & -0.12 & 0.09 \\ 
 & 2 &   1 & x3 & 0.30 & 0.21 & 0.18 \\ 
 & 2 &   2 & \cellcolor{cyan}x8 & 1.00 & -0.11 & 0.15 \\ 
 & 3 &   1 & x8 & 0.23 & -0.08 & 0.13 \\ 
 & 3 &   1 & \cellcolor{pink}x3 & 0.70 & 0.17 & 0.18 \\ 
 & 3 &   1 & x1 & 0.07 & 0.04 & 0.17 \\ 
 & 3 &   2 & x8 & 0.30 & -0.18 & 0.17 \\ 
 & 3 &   2 & \cellcolor{pink}x3 & 0.70 & 0.20 & 0.18 \\ 
 & 3 &   3 & x8 & 0.03 & -0.48 &  \\ 
 & 3 &   3 & \cellcolor{pink}x3 & 0.97 & 0.10 & 0.18 \\ 
 & 3 &   4 & \cellcolor{pink}x3 & 0.97 & 0.08 & 0.23 \\ 
   \hline
\end{tabular}

\end{footnotesize}
\label{tab:robustness}
\end{table}

\subsection{Real-World Examples}
\label{app:titanic}
\paragraph{Titanic dataset}
In Section \ref{sec:real_world}, we applied REPID on the \textit{titanic} dataset \citep{dawson:1995}. The labeled part of the dataset consists of 11 features and the binary survival target variable of 891 passengers. The features of the raw dataset include: \textit{PassengerId , Name, Pclass, Sex, Age, SibSp, Parch, Ticket, Fare, Cabin, Embarked}, a detailed definition of each feature can be found at \url{https://www.kaggle.com/c/titanic/data}. To fit the RF model and analyze the predictions, we first pre-processed the data according to the following kaggle notebook \url{https://www.kaggle.com/nitinar1/titanic-solution-using-random-forest-tool-r}. The pre-processing steps can be summarized as follows:
\begin{itemize}
    \item[1] We extract a title from the feature \textit{Name} and categorize them into 5 categories (Master, Miss, Mr, Mrs and Rare Title).
    \item[2] We create a family size feature \textit{FsizeD} from the features \textit{Sibsp} as the number of siblings and \textit{Parch} as the number of parents and children, and we categorize it into singleton, small and large family size.
    \item[3] We impute missing values of feature \textit{Embarked} based on the fare price they paid.
    \item[4] We impute missing values of feature \textit{Fare} by its median value of the respective \textit{Pclass} and \textit{Embarked} categories.
    \item[5] We impute the feature \textit{Age} using a random forest imputation via multivariate imputation by chained equations.
    \item[6] We exclude the features \textit{PassengerId, Name, Ticket, Cabin} from the dataset, which leaves us with nine features: \textit{Pclass, Sex, Age, SibSp, Parch, Fare, Embarked, Title, FsizeD}.
\end{itemize}

\begin{figure}[tbh]
    \centering
    \includegraphics[width=0.6\linewidth]{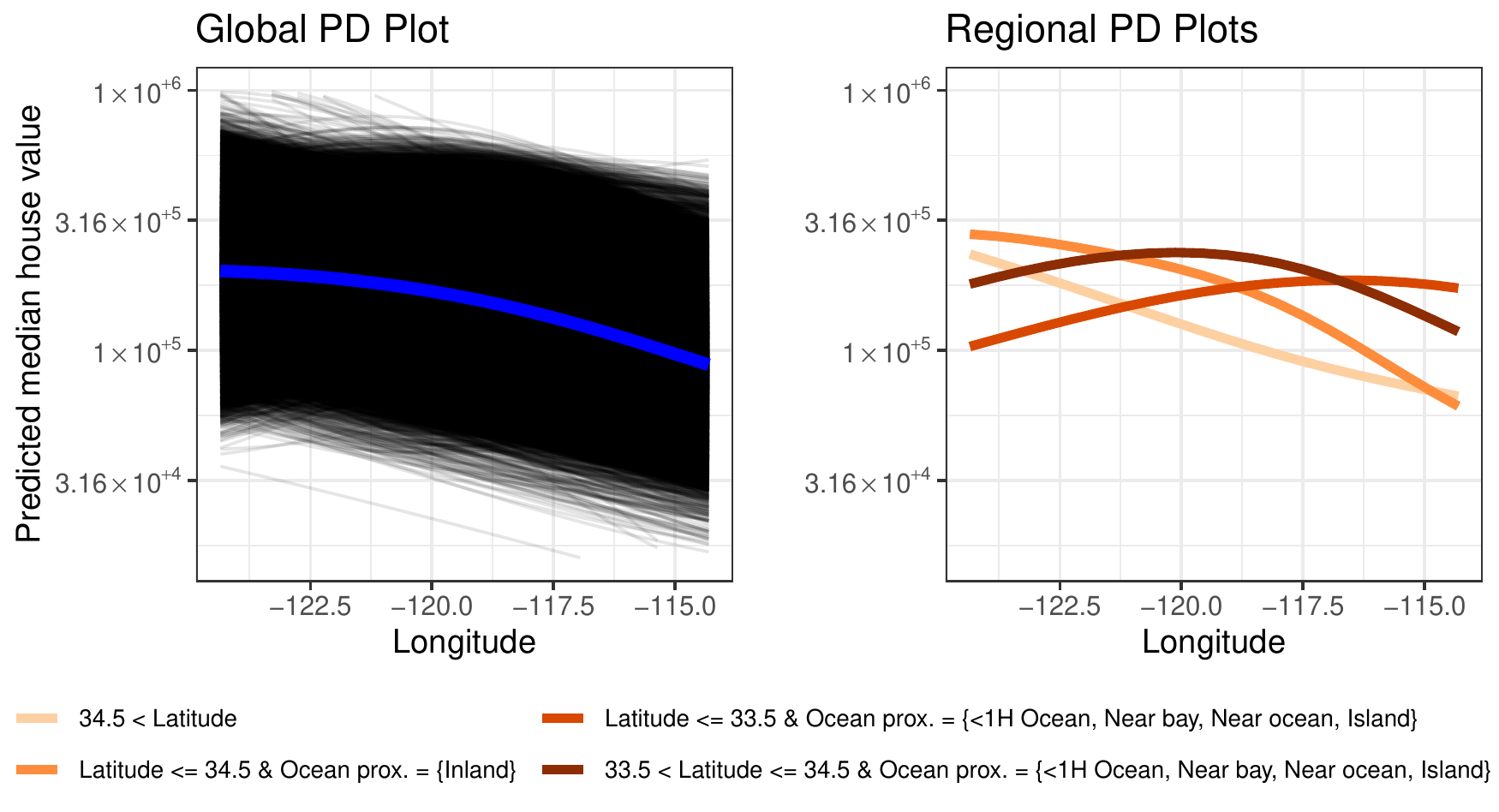}
    \vspace{.2in}
    \caption{The figure shows the global PD plot (blue), including ICE curves (left) and the REPs after applying REPID (right) for the feature of interest Longitude of the \textit{California housing} dataset. The interaction importance $intImp_j$ between Longitude and the interacting features is $0.49$ (Latitude), $0.18$ (Ocean proximity), and $R^2_{int} = 0.67$.}
    \label{fig:california_housing_longitude}
\end{figure}

\paragraph{California housing dataset}
As a second example, we applied REPID 
on the \textit{California housing} dataset \citep{pace1997sparse}. The dataset contains information from the 1990 U.S. Census in California. Each of the 20640 observations provides information of a block group (small geographical unit), with an average population of around 1425 on the median house value (target), eight numeric features, and one categorical feature describing the ocean proximity. The features of the dataset include: \textit{Longitude, Latitude, Housing median age, Total rooms, Total Bedrooms, Population, Households, Median Income and Ocean proximity}. A detailed definition of each feature can be found at \url{https://www.kaggle.com/camnugent/california-housing-prices}.
Only the feature \textit{Total bedroom} contains 207 missing values, which we imputed by the median value of \textit{Total bedroom} of all other observations.
Before applying the neural network on the data, we log transformed the target variable with a base of 10 and log transformed the features \textit{Total rooms, Total Bedrooms, Population, Households, Median Income} using the natural logarithm.
After pre-processing the data, we fit a neural net with one hidden layer of size $20$, a weight decay of $0.1$, and a maximum number of iterations of $1000$. Thus, we obtain a mean absolute error (R-squared) of $0.08$ ($0.78$) under 5-fold cross-validation.
The left plot in Figure \ref{fig:california_housing_longitude} shows that the median house value on average decreases the farther west a house is. The effect of individual observations seems to vary. However, visualizing ICE curves for such a high number of observations is not very insightful. In the right plot, we illustrate the resulting REPs after applying REPID with the same configurations as used for the titanic example in Section \ref{sec:real_world} but with $\gamma = 0.25$. The REPs show that the marginal effect of Longitude on the predicted median house value highly depends on how far north a house is (Latitude: the higher the value the farther north) and how close the house is to the ocean (Ocean proximity). For example, median values of houses that are farther north decrease with Longitude (light orange), while median values of houses farther south and not in the inland increase with Longitude (red).

\begin{figure}[tbh]
    \centering
    \includegraphics[width=0.6\linewidth]{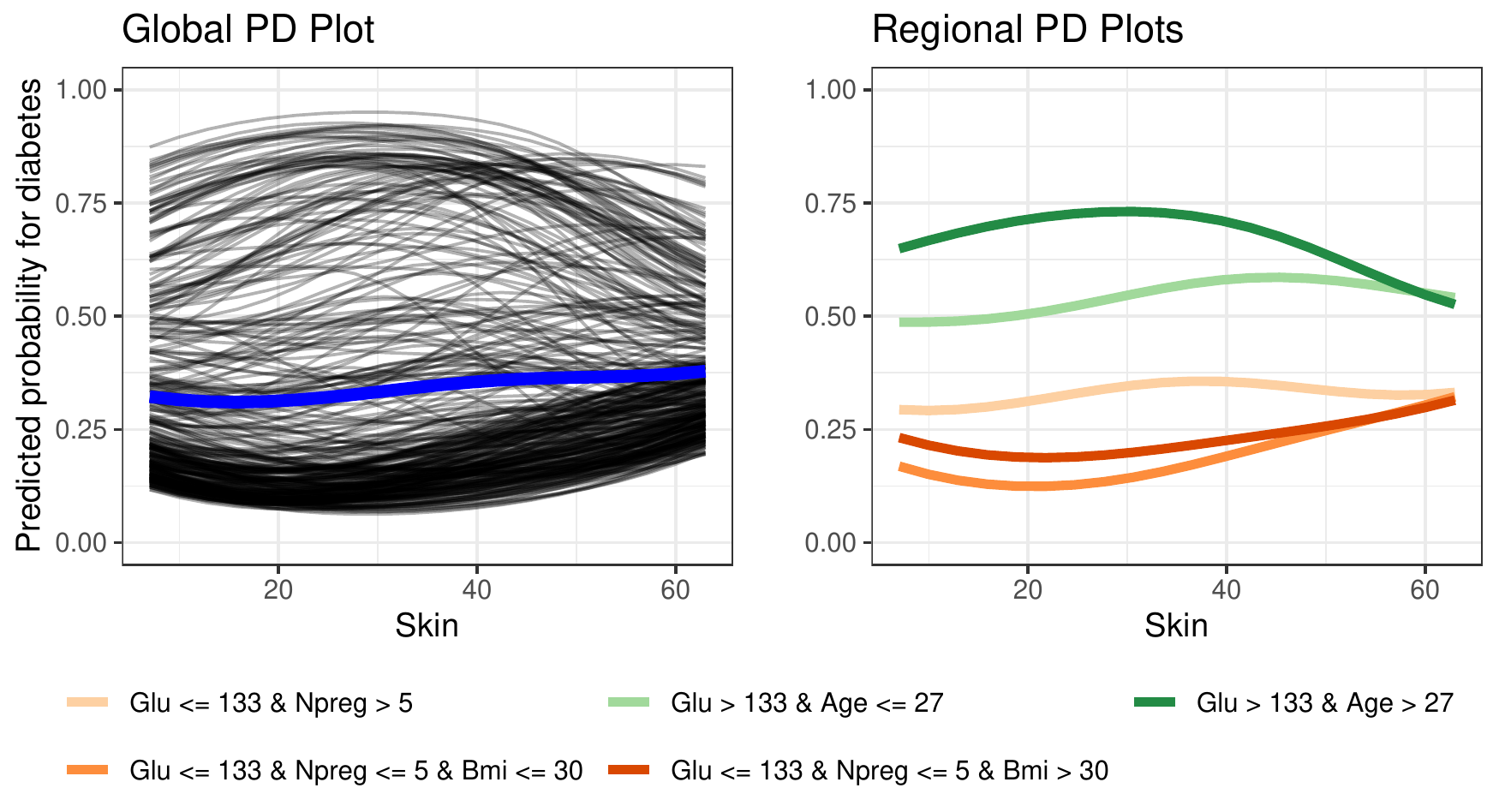}
    \vspace{.2in}
    \caption{The figure shows the global PD plot (blue), including ICE curves (left), and the REPs after applying REPID (right) for the feature of interest Skin of the \textit{diabetes} dataset. The interaction importance $intImp_j$ between Skin and the interacting features is $0.29$ (Glu), $0.09$ (Age), $0.08$ (Npreg), $0.03$ (Bmi) and $R^2_{int} = 0.49$.}
    \label{fig:pima}
\end{figure}

\paragraph{Diabetes dataset}
As a third real-world example, we apply REPID on the \textit{Diabetes dataset}, which analyzes diabetes in Pima Indian women and is available in the \textit{MASS} package in R. The dataset consists of seven numeric features and the binary target variable \textit{type}, which indicates if a woman is diabetic. The features for the 332 women contained in the dataset include: \textit{Npreg} (number of pregnancies), \textit{Glu} (plasma glucose concentration), \textit{Bp} (diastolic blood pressure in mm Hg), \textit{Skin} (triceps skin fold thickness in mm), \textit{Bmi} (body mass index, \textit{ped} (diabetes pedigree function), \textit{Age}. We trained an SVM using 
epsilon support vector regression with a Gauss kernel, $C = 1$ and $\epsilon = 0.1$. Subsequently, we obtained a balanced accuracy of 0.72 using a 5-fold cross-validation.
We are interested in how the feature \textit{Skin} influences the predicted probability for diabetes. When looking at the global PDP in Figure \ref{fig:pima}, one would assume that the skin fold thickness does not effect the predicted probability for diabetes, however, the ICE curves in the left plot indicate heterogeneous effects and, hence, interactions. We apply REPID with the same configurations as used in the \textit{titanic} example in Section \ref{sec:real_world} and obtain the REPs shown in the right plot of Figure \ref{fig:pima}. While the risk of diabetes is in general higher for women with a glucose concentration higher than 133 than for women with a lower glucose concentration, the REPs also show that the risk for women with high glucose concentration values first increases with skin fold thickness and then decreases (green and light green curves), while the risk of diabetes for women with lower glucose concentration values and a maximum of five pregnancies first slightly decreases until a thickness of approximately 20 mm and then increases with skin fold thickness (orange and red curve).  

\end{document}